\documentclass{article}

\usepackage{PRIMEarxiv}
\usepackage[utf8]{inputenc} 
\usepackage[T1]{fontenc}    
\usepackage{hyperref}       
\usepackage{url}            
\usepackage{booktabs}       
\usepackage{amsfonts}       
\usepackage{nicefrac}       
\usepackage{microtype}      
\usepackage{lipsum}
\usepackage{fancyhdr}       
\usepackage{graphicx}       
\usepackage{listings}
\usepackage{mdframed}
\usepackage{xcolor}
\usepackage[round]{natbib}
\usepackage{comment}
\usepackage{tabularx}
\usepackage{caption}
\usepackage{float}
\graphicspath{{img/}}     
\usepackage{booktabs} 
\usepackage{array} 
\newfloat{listing}{htbp}{loa}  
\floatname{listing}{Listing} 

\title{RAG vs Fine-tuning: Pipelines, Tradeoffs, and a Case Study on Agriculture}

\author{
   \\
  \textbf{Microsoft}\\
Angels Balaguer, 
Vinamra Benara, 
Renato Cunha, 
Roberto Estevão, 
Todd Hendry, 
Daniel Holstein, \\
Jennifer Marsman, 
Nick Mecklenburg, 
Sara Malvar, 
Leonardo O. Nunes, 
Rafael Padilha, 
Morris Sharp, \\
Bruno Silva, 
Swati Sharma, 
Vijay Aski, 
Ranveer Chandra
}

\begin{document}
\maketitle

\begin{abstract} 
There are two common ways in which developers are incorporating proprietary and domain-specific data when building applications of Large Language Models (LLMs): Retrieval-Augmented Generation (RAG) and Fine-Tuning. RAG augments the prompt with the external data, while fine-Tuning incorporates the additional knowledge into the model itself. However, the pros and cons of both approaches are not well understood. In this paper, we propose a pipeline for fine-tuning and RAG, and present the tradeoffs of both for multiple popular LLMs, including Llama2-13B, GPT-3.5, and GPT-4. Our pipeline consists of multiple stages, including extracting information from PDFs, generating questions and answers, using them for fine-tuning, and leveraging GPT-4 for evaluating the results. We propose metrics to assess the performance of different stages of the RAG and fine-Tuning pipeline. We conduct an in-depth study on an agricultural dataset. Agriculture as an industry has not seen much penetration of AI, and we study a potentially disruptive application - what if we could provide location-specific insights to a farmer? Our results show the effectiveness of our dataset generation pipeline in capturing geographic-specific knowledge, and the quantitative and qualitative benefits of RAG and fine-tuning. We see an accuracy increase of over 6 p.p. when fine-tuning the model and this is cumulative with RAG, which increases accuracy by 5 p.p. further. In one particular experiment, we also demonstrate that the fine-tuned model leverages information from across geographies to answer specific questions, increasing answer similarity from 47\% to 72\%. Overall, the results point to how systems built using LLMs can be adapted to respond and incorporate knowledge across a dimension that is critical for a specific industry, paving the way for further applications of LLMs in other industrial domains.

\end{abstract}

\keywords{GPT-4 \and Agriculture \and Retrieval Augmented Generation \and Fine-tuning}

\section{Introduction}

Over the past few years, artificial intelligence and natural language processing have seen significant advancements, leading to the development of powerful large language models (LLMs) such as the Generative Pre-trained Transformer (GPT). The technology driving LLMs, including advanced deep learning techniques, large-scale transformers, and vast amounts of data, have propelled their rapid evolution. Models like GPT-4~\citep{OpenAI2023} and Llama 2~\citep{touvron_llama_2023-1} have demonstrated exceptional performance across numerous tasks and domains, often without specific prompts. These models surpass their predecessors and hold immense potential in various fields like coding, medicine, law, agriculture, and psychology, closely approaching human-level expertise~\citep{bubeck2023sparks, nori2023capabilities, demszky2023using}. As LLM research continues, it is critical to identify their limitations and address the challenges of developing more comprehensive artificial general intelligence (AGI) systems. Moreover, the machine learning community must move beyond traditional benchmarking datasets and evaluate LLMs in ways that closely resemble human cognitive ability assessments.

The adoption of Artificial Intelligence (AI) copilots across various industries is revolutionizing the way businesses operate and interact with their environment. These AI copilots, powered by LLMs, provide invaluable assistance in data processing and decision-making processes. In healthcare, for example, AI copilots are being leveraged to predict patient risks and improve diagnostic accuracy~\citep{kim2023, thirunavukarasu2023,Alowais2023}. In manufacturing, they aid in enhancing operational efficiency, reducing downtime, and improving product quality~\citep{vanti2023llm,li2023large}. In the realm of finance, AI copilots help in fraud detection, risk management, and investment decision-making~\citep{AI4Finance-Foundation2022,venasolutions2022}. By harnessing the power of AI copilots, industries can drive innovation, optimize performance and gain a competitive edge.

Despite these advancements, the application of AI in specific fields such as agriculture is still limited due to a lack of specialized training data. While AI has been used to derive insights from satellite imagery and sensor data in agriculture~\citep{farmbeats, farmbeats1, farmvibes, SpaceEye, deepmc, sharma2023knowledge}, the technology is still slowly being adopted by farmers.
While GPT-4 and Bing are powerful tools for finding information, they may not provide the best solutions for farmers who have very specific questions about their crops and livestock. These questions often require knowledge of local conditions, specific varieties, and up-to-date data that might not be readily available through general search engines~\citep{rfi}. As an example, Table~\ref{table:bingexamples} compares the answers from GPT-4 and an agronomist expert to the same query asked for three different U.S. states. While an expert would provide contextualized answers grounded on the states specific climate and agriculture tradition, LLMs provide a generic answer that, although correct, is not as precise for each state as the expert answer.
\renewcommand{\arraystretch}{1.5}
\begin{table}
\centering    
\begin{tabular}{p{3cm}p{12cm}}    
\hline
\textbf{State} & \textbf{What is the best times to plant trees and shrubs in \rule{2cm}{0.4pt} ?} \\
\hline
\textit{GPT-4 answers} & \\
Arkansas & The best time to plant trees and shrubs in Arkansas is during the \textcolor[HTML]{239f3f}{\textbf{fall and early spring seasons}}. Planting in the fall (October to November) allows the roots to establish themselves before the hot summer temperatures arrive. Early spring (February to March) is also a good time to plant, as it gives the plants a chance to grow and establish themselves before the summer heat sets in.\\ 
Connecticut & The best time to plant trees and shrubs in Connecticut is during the \textcolor[HTML]{f7485c}{\textbf{early spring and fall seasons}}. Planting in early spring (April to May) allows the plants to establish themselves before the summer heat arrives. Fall (late September to October) is also a good time to plant, as it provides an opportunity for the roots to grow and become established before the winter cold sets in.\\ 
Georgia &  The best time to plant trees and shrubs in Georgia is during the \textcolor[HTML]{b1b511}{\textbf{fall and early spring seasons}}. Planting in the fall (October to November) allows the roots to establish themselves before the hot summer temperatures arrive. Early spring (February to March) is also a good time to plant, as it gives the plants a chance to grow and establish themselves before the summer heat sets in.\\ \hline

\hline
\textit{Bing Chat answers} & \\
Arkansas & The best time to plant trees and shrubs in Arkansas is generally in the \textcolor[HTML]{239f3f}{\textbf{fall}}. More specifically, September through October is often recommended. This allows enough time for the plants to establish strong roots before the ground starts to freeze. However, \textcolor[HTML]{239f3f}{\textbf{spring}} can also be a good time to plant when soil temperatures are mild. \\ 
Connecticut &  The best time of the year to plant most trees and shrubs is in the \textcolor[HTML]{f7485c}{\textbf{fall}}. The cooler temperatures are easier on the trees and shrubs than the harsh summer heat, and the autumn rain helps them establish a stronger root system.\\ 
Georgia & The best time to plant trees and shrubs in Georgia is during the \textcolor[HTML]{b1b511}{\textbf{fall and winter months}}. Trees planted in November and December have time to produce new roots before having to supply water to lots of thirsty leaves next spring \\ \hline

\textit{Expert's answers} & \\
Arkansas & The best times to plant trees and shrubs are in the \textcolor[HTML]{239f3f}{\textbf{spring and fall}}. Spring planting is recommended because the soil is typically moist and warm, allowing for easy digging and root growth.\\ 
Connecticut & The best time to plant shrubs in shady areas is during \textcolor[HTML]{f7485c}{\textbf{late winter or early spring}} in well-prepared soil.\\  
Georgia & The best time to plant trees and shrubs is in the \textcolor[HTML]{b1b511}{\textbf{fall, specifically during the dormant season,}} which is when the plant has just entered dormancy or is in full dormancy.\\\hline    
\end{tabular}\vspace{0.2cm}    
\caption{Comparison of GPT-4, Bing Chat, and expert's answers for an example query (\textit{``What is the best times to plant tress and shrubs?''}), considering three locations (\textit{Arkansas}, \textit{Connecticut}, and \textit{Georgia}). GPT-4 is not able to incorporate location-specific knowledge, providing the same answer in all cases without considering geographical particularities. Enhanced by a search engine, Bing Chat does a better job adapting the answer to each location, but the expert's answers are still more precise.}    
\label{table:bingexamples}    
\end{table}

In this paper, we introduce a new focus: the creation of AI copilots for industries that require specific contexts and adaptive responses, such as the agriculture industry. We propose a comprehensive LLM pipeline to generate high-quality, industry-specific questions and answers. This approach involves a systematic process comprising the identification and collection of relevant documents encompassing a wide range of agricultural topics. These documents are then cleaned and structured to facilitate the generation of meaningful Q\&A pairs using the base GPT model. The generated pairs are subsequently evaluated and filtered based on their quality. Our goal is to create a valuable knowledge resource for an specific industry, with a case study in agriculture with the goal of ultimately contributing to the advancement of this crucial field.

The proposed pipeline aims to generate domain-specific questions and answers catering to professionals and stakeholders in an industry where answers from a copilot are expected to be grounded by relevant industry-specific factors. In the case of our agriculture study, we are aiming to generate geography-specific answers. For this, our starting point is an agriculture dataset, which is fed into three main components: Q\&A generation, Retrieval-Augmented Generation (RAG)~\citep{lewis2020retrieval}, and fine-tuning process. The Q\&A generation creates question and answer pairs based on the information available in the agriculture dataset, while RAG uses it as a knowledge source. The generated data is then refined and used to fine-tune several models while their quality is evaluated using a combination of proposed metrics. Through this comprehensive approach, we aim to harness the power of LLMs for the benefit of the agriculture industry and its stakeholders.

In this paper, we make several noteworthy contributions to the understanding of LLMs in the agriculture domain. These contributions can be itemized as follows:

\begin{itemize}
\item \textbf{Comprehensive evaluation of LLMs:} we conducted an extensive evaluation of large language models, including LlaMa2-13B, GPT-4 and Vicuna~\citep{zheng_judging_2023}, in answering agriculture-related questions. This was done using benchmark datasets from major agriculture producer countries. Our evaluation included the complete fine-tuning and RAG pipeline, each with its own set of metrics. The findings from this evaluation provide a crucial baseline understanding of the performance of these models within the agricultural context. Moreover, we conduct evaluations that demonstrate the impact of spatial shift on the knowledge encoded by existing LMs and the improvements offered by spatially-scoped fine-tuning. In our analysis, GPT-4 consistently outperformed other models, however the cost associated with its fine-tuning and inference needs to be taken into consideration.
\item \textbf{Impact of retrieval techniques and fine-tuning:} we investigated the impact of retrieval techniques and fine-tuning on the performance of LLMs. The study revealed that both RAG and fine-tuning are effective techniques for improving the performance of LLMs. RAG was shown to be highly effective in instances where data is contextually relevant, such as in the interpretation of farm data, while also leading to more succinct responses than the base model. Fine-tuning, on the other hand, was found to be useful in teaching the model new skills specific to the agricultural domain, and providing more precise and succinct responses. However, the high initial cost due to the extensive work required to fine-tune the model on new data is a significant consideration.
\item \textbf{Implications for potential uses of LLMs in different industries:} this research serves as a pioneering step in establishing a pipeline for the application of RAG and fine-tuning techniques in LLMs, fostering innovation and collaboration across multiple industries. With our initial focus on agriculture, we demonstrated how these strategies can lead to more efficient models starting from the Q\&A generation process. The insights gained from this study could be applied to other sectors, potentially leading to the development of more efficient AI models for a variety of applications. For instance, one potential application could be in the development of AI copilots for different industries, where the ability to provide accurate, relevant and succinct responses to user queries is paramount.
\end{itemize}

The remainder of this work is organized as follows. Section~\ref{sec:methodology} presents the methodology in detail, including the data acquisition process, information extraction procedure, question and answer generation, and fine-tuning of the model. We then describe in Section~\ref{sec:dataset} the dataset used in the study, which includes data from the USA, Brazil, and India. In Section~\ref{sec:metrics}, we outline the metrics used to evaluate the effectiveness of the proposed methodology, focusing on both question and answer evaluation. Section~\ref{sec:experiments} presents a comprehensive evaluation of various models and their performance in generating question-answer pairs within the context of agricultural data, using these for RAG on GPT-4, Vicuna~\citep{zheng_judging_2023} and Llama2 13B~\citep{touvron_llama_2023, touvron_llama_2023-1} and fine-tuning of GPT-4 and Llama2 13B. Finally, we conclude with a summary of the main findings and possible directions for future research in Section~\ref{sec:conclusion}.

\section{Methodology}\label{sec:methodology}

\begin{figure}
  \centering
  \includegraphics[width=0.9\textwidth]{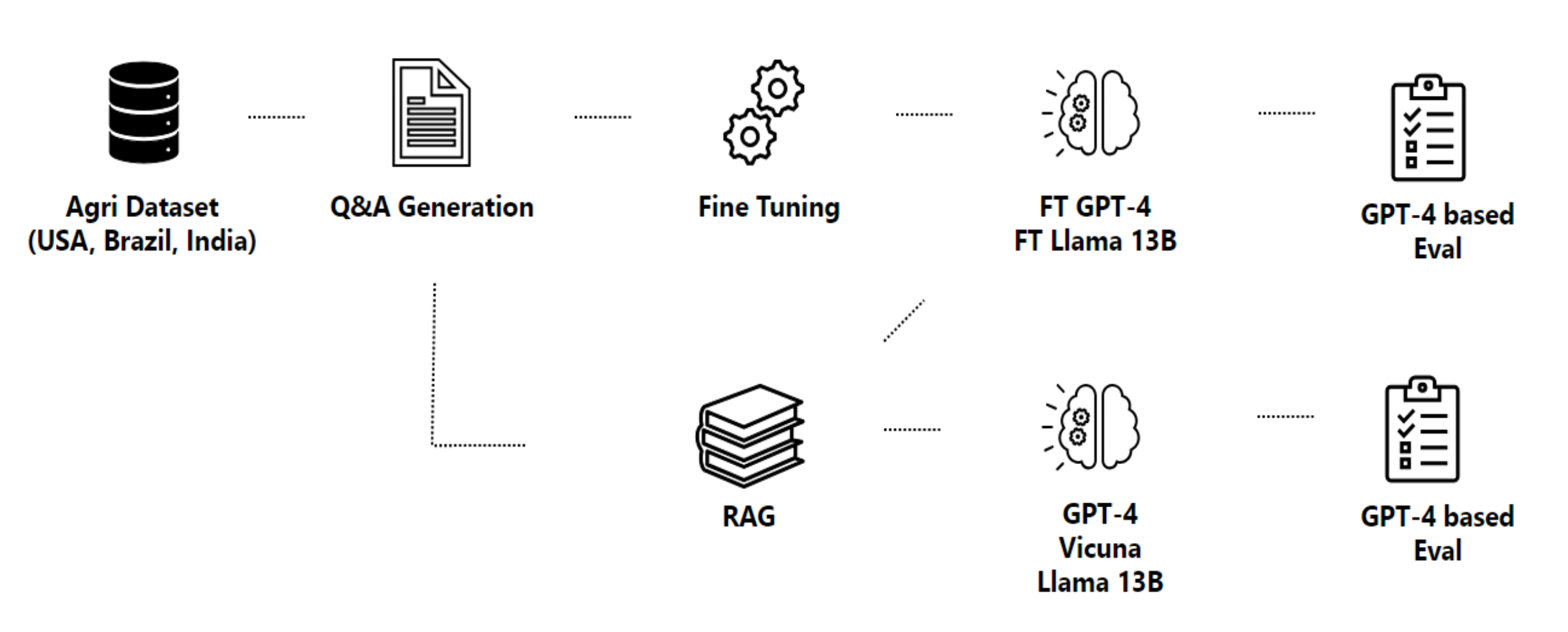}
  \caption{Methodology pipeline. Domain-specific datasets are collected, and the content and structure of the documents are extracted. This information is then fed to the Q\&A generation step. Synthesized question-answer pairs are used to fine-tune the LLMs. Models are evaluated with and without RAG under different GPT-4-based metrics.}
  \label{fig:diagram}
\end{figure}

The methodology proposed in this paper revolves around a pipeline designed to generate and evaluate question-answer pairs for building domain-specific copilots. The proposed pipeline is shown in Figure~\ref{fig:diagram}.

The pipeline begins with data acquisition, described in Section~\ref{sec:dataacquisition}. The initial focus is on gathering a diverse and curated dataset pertinent to the industry domain. This includes sourcing data from various high-quality repositories such as government agencies, scientific knowledge databases, and proprietary data, if needed. The details of potential data sources and the types of documents selected are exemplified and further elaborated in Section~\ref{sec:dataset}.

Following data acquisition, the pipeline proceeds to extract information from the collected documents. This step is crucial as it involves parsing complex and unstructured PDF files to recover the underlying content and structure. This process, detailed further in Section~\ref{pdfextraction}, employs robust text extraction tools and machine learning algorithms to recover textual, tabular, and visual information, while also identifying the semantic structure of the documents and possible cross-relations in them. 

The next component of the pipeline is question and answer generation. The objective here is to generate contextually grounded and high-quality questions that accurately reflect the content of the extracted text. The methodology employs a framework to control the structural composition of both inputs and outputs, thereby augmenting the overall efficacy of response generation from language models. This part of the process is detailed in Section~\ref{sec:augmentation}.

Subsequently, the pipeline generates answers for the formulated questions. The methodology employed here leverages Retrieval-Augmented Generation (RAG), which combines the power of retrieval and generation mechanisms to create high-quality answers. The answer generation process is discussed further in Section~\ref{sec:methodology_answer_generation_rag}.

Finally, the pipeline fine-tunes the models with the Q\&A pairs. The optimization process, discussed in the Section~\ref{ft}, employs methods like Low Rank Adaptation (LoRA)~\citep{hu2021lora} and ensures a comprehensive understanding of the content and context of the scientific literature, making it a valuable resource for various domains or industries.

In the following sections, we will delve deeper in each components of the pipeline, highlighting their objectives, inputs and outputs, and reasoning behind why they were added to the pipeline.

\subsection{Data Acquisition}\label{sec:dataacquisition}

The initial focus of the pipeline is to gather an assorted and well-curated dataset that captures information of interest to an industry. This enables the generation of questions and answers, forming the foundation for refining the models to produce more precise and pertinent responses. For this step, we seek data sources that contain high-quality, authoritative information on the topic of interest. For example, in agriculture, this includes agricultural and environmental government agencies, scientific knowledge repositories, and agronomist exams databases. It is also important that the information extracted to be aligned with the grounding that will be provided to the model. For instance, in the case of agricultural data, we sourced guidelines and procedures that were geography-specific, i.e. with a shared location among documents. We provide further details on the sources and type of documents selected for this step in Section~\ref{sec:dataset}. 

With authoritative sources defined, web scraping tools come into play to gather the required data. We employed web scraping frameworks, including Scrapy~\citep{scrapy} and BeautifulSoup~\citep{beautifulsoup}, to parse through the websites, uncovering all available documents and downloading the relevant files. 

\subsection{PDF Information Extraction}\label{pdfextraction}

\begin{figure}
  \centering
  \includegraphics[width=0.95\textwidth]{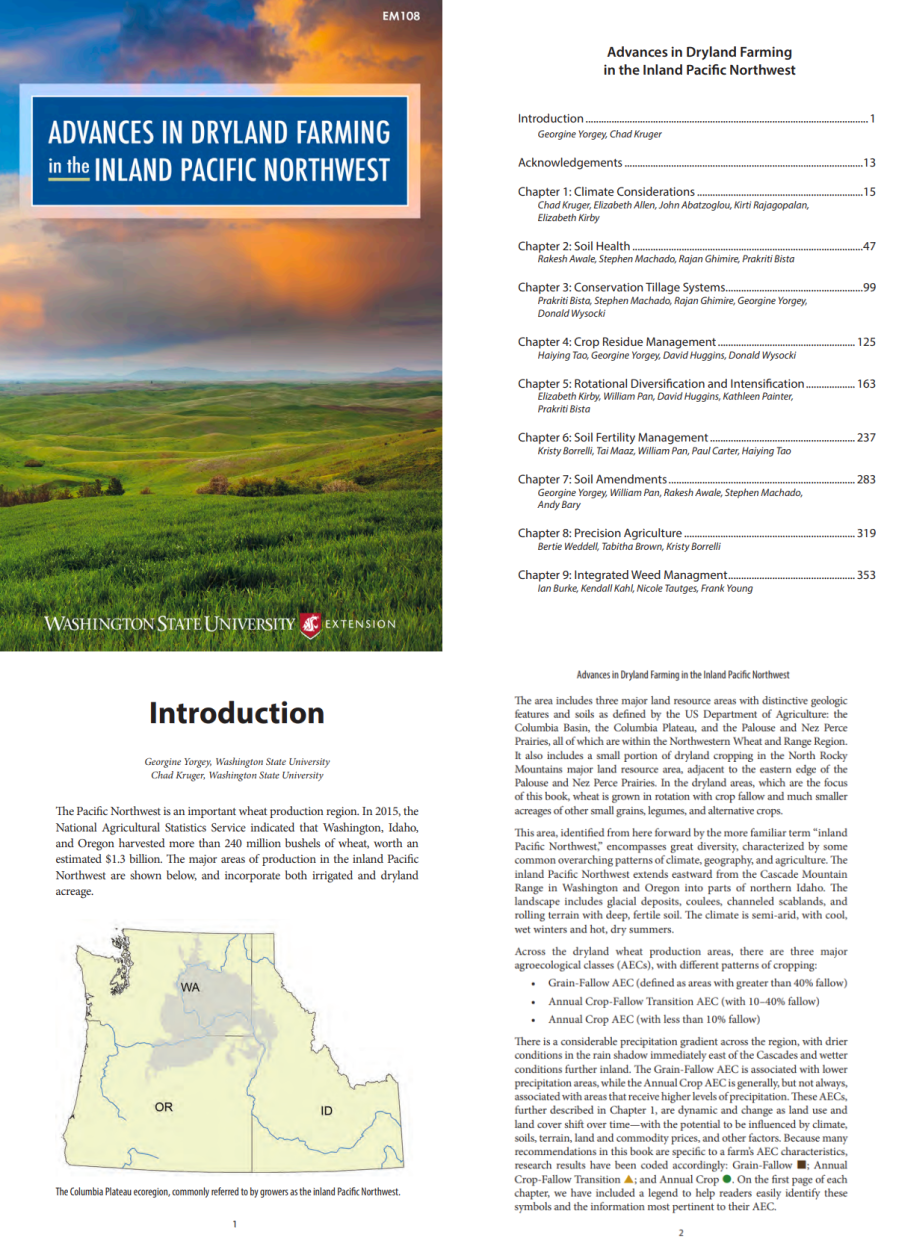}
  \caption{Example of a document from Washington state present in our dataset. The diverse layouts of PDF files, which often include textual and visual data, pose a significant challenge in terms of extracting not just the content, but also the underlying structure.}
  \label{fig:pdf_file_example}
\end{figure}

In our study, the extraction of information and text structure from the collected documents is critical to the quality of the subsequent steps. However, this is a challenging task as the primary purpose of PDFs is to accurately display a document across different systems, and not for easy information extraction. The underlying structure of a PDF file does not map onto the logical structure of a document, i.e., sections, subsection, and associated content. Additionally, with documents originating from various sources, we observe their layouts and formatting to be complex and lack standardization, often presenting a mixture of tables, images, sidebars, and page footers. We present in Figure~\ref{fig:pdf_file_example} an example of a PDF file in our dataset.   

With this in mind, the main objective of this step of the pipeline is to address the complexities inherent in processing data derived from a range of formatted PDF documents. This is achieved by leveraging robust text extraction tools and machine learning algorithms that employ advanced natural language processing techniques. The focus is not only in recovering the content of each file, but also its structure. Among other things, we are interested in discovering what are the sections and subsections, parsing the information presented in tables and diagrams, identifying cross-references within the document, and linking images with their caption and description. By retrieving the organization of the document, we can easily group information, reason over numerical data present in tables, and provide more consistent snippets of the text to the Q\&A generation step. It is also very important that all available information is extracted from the document, with well-formed sentences.

There are multiple tools available online that extract information from PDFs~\citep{pdf2text, pypdf}. However, many of them lack the ability to retrieve content in a structured way. For example, \textit{pdf2text} is an open-source Python library offering methods to iterate over PDF's pages and recover the textual information. We provide in Listing~\ref{listing:pdf2text} the output of \textit{pdf2text} over the document from Figure~\ref{fig:pdf_file_example}. The library is able to recover the textual information, but markers representing the beginning of a section or subsection are lost within the retrieved data, hindering our ability to reason over the document structure. Captions of tables and figures are also lost in conversion but sometimes contain critical information for the understanding of the document.

\begin{listing}
    \begin{mdframed}[linewidth=1pt,linecolor=black,backgroundcolor=gray!20]
  \centering
\begin{lstlisting}
Advances in Dryland 
Farming in the Inland 
Pacific Northwest
Georgine Yorgey and Chad Kruger, editorsFor Sanford Eigenbrode, in recognition of his resolute effort
leading the REACCH project.
For Iris, Toby, Leah, Jocelyn, Alexis, and Zakkary, who symbolize 
why this work is so important.College of Agricultural, Human, and Natural Resource Sciences
Use pesticides with care. Apply them only to plants, animals, or sites as listed on the label. When 
mixing and applying pesticides, follow all label precautions to protect yourself and others around you. It is a violation of the law to disregard label directions. If pesticides are spilled on skin or clothing, remove clothing and wash skin thoroughly. Store pesticides in their original containers and keep them out of the reach of children, pets, and livestock.
2017 Washington State University WSU Extension bulletins contain material written and produced for public distribution. Alternate 
formats of our educational materials are available upon request for persons with disabilities. Please contact Washington State University Extension for more information.
Y ou may order copies of this and other publications from WSU Extension at 1-800-723-1763 or 
http://pubs.wsu.edu.

...

5871Introduction
Georgine Yorgey, Washington State University
Chad Kruger, Washington State University
The Pacific Northwest is an important wheat production region. In 2015, the 
National Agricultural Statistics Service indicated that Washington, Idaho, 
and Oregon harvested more than 240 million bushels of wheat, worth an estimated 1.3 billion. The major areas of production in the inland 
\end{lstlisting}
\end{mdframed}
  \caption{Text extracted from the PDF presented in Figure \ref{fig:pdf_file_example} using PDF2text. Structural markers, such as the \textit{``Introduction''} section header, end up lost within the parsed text.}
  \label{listing:pdf2text}
\end{listing}

Considering this, we employed GROBID (GeneRation Of BIbliographic Data)~\citep{GROBID}, a machine learning library specifically tailored for extracting and processing data from scientific literature in PDF format. The goal is to transform unstructured PDF data into structured data in the form of TEI (Text Encoding Initiative) format~\citep{tei}, efficiently managing large volumes of files. The use of GROBID, trained on a vast corpus of scientific articles, enables the recognition of a wide array of document elements and extraction of associated bibliographic data. We illustrate its capabilities in Listing~\ref{lst:example-json} with the output of GROBID for the document from Figure~\ref{fig:pdf_file_example}.

From the GROBID-generated TEI files, we extracted a subset of the sections of the TEI files comprising the document metadata (title, authors, abstract), sections, tables, figure references, bibliography, and the content itself. Crucially, this phase underscores the belief that the structure of the text is as important as its content. The final objective is to convert the TEI files into more manageable JSON files that preserve not only the content, but also the structure of the original PDFs. This approach ensures a comprehensive understanding of the content and context of the scientific literature, making it a valuable resource for various domains or industries.

\begin{listing}
    \begin{mdframed}[linewidth=1pt,linecolor=black,backgroundcolor=gray!20]
  \centering
\begin{lstlisting}
{
    "grobid_version": "0.7.3",
    "grobid_timestamp": "2023-07-04T13:05+0000",
    "language_code": "en",
    "citations": [
        {
            "authors": [
                {
                    "given_name": "J",
                    "surname": "Abatzoglou",
                    "name": "J T Abatzoglou"
                },
                {
                    "given_name": "T",
                    "surname": "Brown",
                    "name": "T J Brown"
                }
            ],
            "index": 0,
            "id": "b0",
            "date": "2012",
            "title": "A Comparison of Statistical Downscaling Methods Suited for Wildfire Applications",
            "journal": "International Journal of Climatology",
            "volume": "32",
            "issue": "5",
            "pages": "772-780"
            ...
            {
                "title": "Introduction",
                "content": [
                    "Climate has always had a dominant influence on dryland production in this region, shaping crop choices, agronomic management systems, and conservation efforts. Though farmers are already highly skilled managers in the context of variable temperature and precipitation patterns, climate change is expected to add uncertainty, stretching the limits of existing management systems. Projected climate change also brings new urgency to questions of sustainability, as changing seasonal climate patterns may exacerbate conditions that have been historically linked to major soil erosion events in the region.",
                ],
                "refs": []...
            }
        }
}
\end{lstlisting} 
\end{mdframed}
  \caption{JSON file extracted from the PDF of Figure \ref{fig:pdf_file_example}. We use GROBID~\citep{GROBID}, which can extract the content and some of the structure (e.g., section and subsection organization) from the document.}
  \label{lst:example-json}
\end{listing}

\subsection{Question Generation}\label{sec:augmentation}

The initial focus of this section of the pipeline is to manage the inherent complexity and variability of natural language when generating questions from the extracted text. We aim to generate contextually grounded and high-quality questions that accurately reflect the content of the extracted text. For this, we employ the Guidance framework~\citep{Guidance}, whose primary advantage lies in its capacity to provide unparalleled control over the structural composition of both inputs and outputs, thereby augmenting the overall efficacy of response generation from language models. This degree of control results in outputs that are not only more precise, but also exhibit enhanced coherence and contextual relevance. The framework's capability to amalgamate generation, prompting, and logical control into a singular, unified process closely parallels the inherent mechanisms of language model text processing. Moreover, the unique feature of Guidance that enables the direction of language models via context-specific prompts, contributes to a heightened level of semantic relevance in the resultant text. In our case, this ensures the questions will carry semantic relevance to the source text while taking into account the context extracted from the JSON files.

First, we augment the content and structure of available documents by explicitly adding supporting tags from the text. We formulated prompts to extract a list of locations and agronomic topics mentioned in each section of the document (e.g., if that section refers to crops, cattle, or diseases), as exemplified in Listing~\ref{lst:tag_location_prompt}, and task the LLM model to answering them based on the data extracted from the JSON files. The aim is to use of the the additional information, including locations and mentioned topics, to ground the generation process, enhancing the relevance of the questions and guiding the model to cover a wide range of topics and challenges. 

Once obtained, we combine the supporting context and section content, prompting the LLM to generate a set of questions based on them. We provide an example of the question generation prompt in Listing~\ref{lst:question_generation_prompt}. The prompt includes a system preamble that guides the LLM into formulating assessment questions related to industrial topics based on the document content. Whereas the user portion provides few-shot examples of the types of expected questions, as well as the content and supporting context used for generation. With this setup, the LLM generates a set of 5 to 15 questions per section of the document.    

\begin{listing}
    \begin{mdframed}[linewidth=1pt,linecolor=black,backgroundcolor=gray!20]
  \centering
\begin{lstlisting}[language=] 
Answer the following questions with a single Yes or No: 

1. The text mentions a specific location (City/State/Country), yes or no? 

2. The text mentions a specific crop, yes or no?: 

3. The text mentions a specific cattle, yes or no?: 

4. The text mentions a specific disease, yes or no?: 
 
Answer the following questions with a python list, or return an empty python list: 

1. If the text mentioned a location or locations, list them: 

2. If the text mentioned a crop or crops, list them: 

3. If the text mentioned a cattle or cattles, list them: 

4. If the text mentioned a disease or diseases, list them: 
\end{lstlisting}
\end{mdframed}
  \caption{Prompt used to identify supporting context (i.e., list of locations and agronomic topics mentioned) from a document during question generation.}
  \label{lst:tag_location_prompt}
\end{listing}

\begin{listing}
    \begin{mdframed}[linewidth=1pt,linecolor=black,backgroundcolor=gray!20]
  \centering
\begin{lstlisting}[language=]
{{#system~}}
You are an expert in agriculture and you are formulating questions from documents to assess the knowledge of a student about agriculture-related topics.

You have access to the following document metadata encoded as JSON: {{context}}
{{~/system}}

{{#user~}}
An example of expected output follows:

Examples:
{examples}

The document is from {{source}} and has title: {{title}}

Add location (state, country) and crop information to each question, if possible.

Please formulate as many questions as possible to assess knowledge of the text below.

{{section}}

{{~/user}}

{{#assistant~}}
{{gen 'answer' max_tokens=2000}}
{{~/assistant}}
\end{lstlisting}
\end{mdframed}
  \caption{Example of the question generation prompt for crop-related documents. When prompting the model, we replace the \texttt{\{\{context\}\}} tag with the supporting information retrieved in the last step, as well as exchange the \texttt{\{\{source\}\}}, \texttt{\{\{title\}\}}, \texttt{\{\{section\}\}} tags for the document source, title, and the section content, respectively. The prompt was directly related to agriculture in our case, but should be adapted with domain-specific knowledge in other scenarios.}  
  \label{lst:question_generation_prompt}
\end{listing}

\subsection{Answer Generation}\label{sec:methodology_answer_generation_rag}

We employ Retrieval-Augmented Generation (RAG)~\citep{lewis2020retrieval}, which is an innovative approach that combines the power of retrieval and generation mechanisms, to create high-quality answers. RAG is particularly useful when dealing with large and complex datasets, as it can effectively recover relevant information associated with a query and use it to enhance the generation process.

The RAG pipeline begins by retrieving, for a given question, the most relevant documents or passages from our dataset. The retrieval system employs techniques such as BM25, Dense Retrieval~\citep{reimers_sentence-bert_2019, ni_large_2022}, and other advanced retrieval mechanisms. The retrieved documents serve as a knowledge source for the subsequent generation phase. Once the relevant passages have been identified, the generation component comes into play. An LLM takes the question and the retrieved information as inputs and generates a contextually appropriate answer. The generation process is guided by the context provided by the retrieved documents, ensuring that the generated Q\&A pairs are accurate, relevant, and informative. More specifically, the generation process follows three steps:

\begin{itemize}
\item Embedding generation and index construction: we compute embeddings from text chunks extracted from the PDF documents in our dataset, using sentence transformers~\citep{reimers_sentence-bert_2019}. We then used Facebook AI Similarity Search (FAISS)~\citep{johnson2019billion}, a library for efficient indexing and similarity search of vectors, to create a database of the embeddings.

\item Retrieval: given a input question, we compute its embedding and retrieve the relevant chunks from the FAISS database. This was done with the FAISS retrieval tool, \textit{similarity\_search\_with\_score}, which allowed us to perform a similarity search that returned the top text snippets that were related to the question.

\item Answer generation: with the question and the retrieved chunks as input, an LLM model was used to synthesize an answer. Specifically, we provided the retrieved information from the FAISS database to GPT-4 as context within a custom prompt, which allowed the generation of domain-specific answers. The answers were properly formatted in a JSON file along with the associated questions to create the Q\&A pair.

\end{itemize}

\subsection{Fine-tuning}\label{ft}

The development of advanced AI language models involves the creation of datasets and the training of models with varying capacities. In this case, a dataset consisting of questions and corresponding answers was generated from Llama2-13b-chat~\citep{touvron_llama_2023, touvron_llama_2023-1} with RAG~\citep{lewis2020retrieval}. Several base completion models with different sizes were trained, including Open-Llama-3b~\citep{openlm2023openllama}, Llama2-7b, and Llama2-13b. These models were designed to predict answer tokens based on question prompts, while masking the prompt tokens to ensure the models would not predict them.   

In order to optimize the performance of these models, we employed a fine-tuning process using 8 H100 GPUs and PyTorch's fully-sharded data parallelism (FSDP)~\citep{paszke_pytorch_2019}. FSDP allows for the sharding of parameters, optimizer states, and gradients, effectively reducing the memory requirements during training~\citep{rajbhandari_zero_2020}. Each transformer module within the models functions as an FSDP module, with activation checkpointing~\citep{chen_training_2016}. The training process involved feeding each GPU a microbatch of 4 samples and accumulating gradients over 4 microbatches, leading to an effective batch size of 128 samples per training step. The training was conducted for 4 epochs, employing automatic mixed precision~\citep{micikevicius_mixed_2018} (AMP) with BFloat16, the Adam optimizer~\citep{kingma_adam_2017, loshchilov_decoupled_2019}, a base learning rate of 2e-5, and a cosine learning rate scheduler~\citep{loshchilov_sgdr_2017} with a linear warmup~\citep{goyal_accurate_2018} comprising 4\% of the total number of steps. The FastChat~\citep{zheng_judging_2023} training script with flash-attention monkeypatching was used throughout the training process.   

Lastly, we also fine-tuned GPT-4 in this setting. Being larger and more expensive, our goal was to assess if the model would benefit from additional knowledge in comparison to its base training. Due to its complexity and the amount of available data, we used Low Rank Adaptation (LoRA)~\citep{hu2021lora} for the fine-tuning process. This technique provides an efficient way to adapt parameter-heavy models, requiring less memory and computational resources compared to traditional re-training. By tuning a reduced set of parameters in the attention modules of the architecture, it embeds domain-specific knowledge from new data without losing knowledge gained during base training. In our study, optimization was done for 4 epochs, with a batch size of 256 samples, and a base learning rate of 1e-4 that decayed as training progressed. The fine-tuning was carried out on seven nodes, each with eight A100 GPUs, over a total runtime of 1.5 days.

\section{Dataset Overview}\label{sec:dataset}
This study evaluates fine-tuned and RAG-enhanced language models using context-related questions and answers originated datasets from three major crop producer countries: USA, Brazil and India. In our case, we are using agriculture as the industrial setting as an example. Available data varied considerably in format and content, ranging from regulatory documents, scientific reports, agronomic exams, to knowledge databases. In this section, we present each dataset in more detail. 

\subsection{USA}

We collected documents, handbooks, and reports publicly available online from the United States Department of Agriculture (USDA), state agriculture and consumer services agencies, and partners from the Land-Grant Institutions National Program. Available documents contain federal regulatory and policy information surrounding crop and livestock management, information on diseases and best practices, quality assurance and export regulations, details on assistance programs, as well as insurance and pricing guidelines. Collected data totals more than 23k PDF files with over 50M tokens, representing 44 states in the USA. We downloaded and preprocessed these files, extracting the textual information that could be used as input to the Q\&A generation pipeline. To benchmark and evaluate the models, we employed the documents related to the Washington state, which comprises 573 files with over 2M tokens. We present in Listing~\ref{fig:usa_dataset_example} an example of content within these documents.

\begin{listing}
\begin{mdframed}[linewidth=1pt,linecolor=black,backgroundcolor=gray!20]
\centering
\begin{lstlisting}
Title: 2009 Cost Estimates of Establishing and Producing Gala Apples in Washington

Section: Gala Apple Production in Washington

Since the first commercial plantings of Gala apples in the 1980s, the popularity of this cultivar has grown exponentially. Production of Gala apples in the United States was motivated by a desire to create export market opportunities, especially in East Asia, where the variety is very popular (Economic Research Service, 2005). Subsequent imports of lower-priced apples from China and other Asian markets, as well as rapid expansion of acres, forced U.S. apple growers to also market these varieties domestically.

...
\end{lstlisting}
\end{mdframed}
\caption{Example of a passage in one of the Washington state documents, focusing on the production of Gala apples.}
\label{fig:usa_dataset_example}
\end{listing}

\subsection{Brazil}

We used a dataset called "500 Questions 500 Answers - Embrapa/SCT" containing books with questions and answers related to various aspects of crop cultivation and management in Brazil. The questions are text-based inquiries formulated by a diverse group of stakeholders, including producers, farmers, and farming associations, and are accompanied by responses from Embrapa specialists. 

The dataset, "500 Questions 500 Answers - Embrapa/SCT" provides a valuable resource for farmers and agronomists to improve their knowledge of various agricultural topics. The questions in the dataset are not presented in a multiple-choice format, which requires a deeper understanding of the subject matter to generate accurate and relevant answers. Although the questions often have no single correct answer, the dataset encompasses a wide range of topics essential to both groups. The study highlighted the importance of understanding the nuances of agricultural topics and the expertise required to evaluate the relevance and correctness of AI-generated answers. The Embrapa dataset aims to foster sustainable agricultural practices and contribute to the overall knowledge base.

Listing \ref{fig:gall_nematode_example} illustrates an example of an adapted
question from the Embrapa dataset, demonstrating the challenges of verifying the
correctness of answers in this context. Since the dataset's questions are not
multiple-choice and often have no single correct answer, assessing the accuracy
of the models can be subjective.

\begin{listing}
\begin{mdframed}[linewidth=1pt,linecolor=black,backgroundcolor=gray!20]
\centering
\begin{lstlisting}
What are the symptoms and damages caused by the gall nematode in citrus plants?

Answer: The symptoms of the gall nematode in citrus plants include dysfunctionin the roots,
leading to smaller leaf size and marked chlorosis. This resultsin a reduction in production.
The growth of the main root may be paralyzed,causing a proliferation of lateral roots. In
severe infections, the plant maydie before the nematode completes its cycle. However, not all
citrus species are susceptible to this nematode, so it is essential to identify the species
and monitor it in the soil and roots.
\end{lstlisting}
\end{mdframed}
\caption{Example of an adapted question from the Embrapa dataset, focusing on
the symptoms and damages caused by the gall nematode in citrus plants.}
\label{fig:gall_nematode_example}
\end{listing}

\subsection{India}
\paragraph{Farmer advisory (KVK Q\&A portal)}: 
The Krishi Vigyan Kendra (KVK) is a specialized agricultural extension center in India, dedicated to the practical application of agricultural research in a local context. Its main objective is to bridge the gap between research and practice. To support farmers, KVKs offer a variety of services, including a helpline to address farmers' questions. This data, which captures these interactions is available here: \url{https://data.gov.in/datasets_webservices/datasets/6622307}. We sourced data from 27 Indian states containing a total of $100,000$ farmer queries between the years $2015$ to $2022$. Each record contains the following fields: growing season (kharif/rabi), crop type, query type (weather related, government schemes, plant protection, seeds, agriculture mechanization), farmer query, answer from the KVK, state and district name and, a timestamp. It's worth noting that both the questions posed by farmers and the responses provided by KVK are originally in the respective local state languages. We used the Azure translation services to translate to English. 
\paragraph{\textbf{Vikaspedia}}: 
Vikaspedia \url{https://vikaspedia.in/agriculture/} portal provides general information on various aspects of agriculture and related topics. There is information on topics such as crop production, agricultural inputs, agricultural exports, agricultural insurance, agro- enterprises, best practices, fisheries, forestry, organic farming, policies and schemes, post harvest technologies, poultry, weather information, and more. It is a useful source of knowledge for farmers, students, researchers, and anyone interested in agriculture.

Listing ~\ref{fig:india-example} an example question and answer from the India dataset. The answers from KVK are brief and consist of at most one sentence. Using the Vikaspedia data, we enhanced the answers from KVK to make them more detailed and clear for the farmers. 
\begin{listing}
\begin{mdframed}[linewidth=1pt,linecolor=black,backgroundcolor=gray!20]
\centering
\begin{lstlisting}
Question: I'm a farmer from Bikaner, Rajasthan growing Citrus fruits and I want to know about the nutriment management for Citrus. Can you tell me how to manage it?

Answer from the Indian farmer advisory: You can manage nutriment in Citrus by spraying Borex. Use 4 gm per liter of water.

RAG Augmented Answer: Considering your location in Bikaner, Rajasthan, appropriate nutrient management in Citrus fruits is crucial for good yield. Following are the practices to manage nutrient in Citrus: 
1. Firstly, ensure that your soil is well-drained and of shallow or medium depth. Deep heavy soils with more than 60% clay content are not suitable for Citrus cultivation. 
2. Secondly, it's beneficial to incorporate organic matter into the soil. Organic matter, like goat manure or farmyard manure, used in your cultivation process can provide necessary nutrients for Citrus trees. 
3. The role of micronutrients is also quite significant in citrus health and productivity. Boron deficiency in citrus causes dieback of feeder roots, gumming and fruit cracking. Application of Borax can effectively treat boron deficiency. Take a solution of Borax and mix it with water at a ratio of about 4 grams per liter of water. 
4. Periodic spraying of this Borax solution will aid in nutrient absorption by the citrus plants and provide effective results. 
5. Careful monitoring of the health of your plants can guide you in knowing when to apply additional nutrient sprays and treatments. Please note that nutrient management may differ according to the specific variety of Citrus you are growing, so it's recommended to consult with local agricultural institutions or experts when developing a comprehensive nutrient management plan. Also, overuse or misuse of micronutrients could harm citrus trees and the environment, so nutrient applications should always be based on a reliable soil and plant analysis.
\end{lstlisting}
\end{mdframed}
\caption{Example of an actual farmer question and answer from the KVK farmer advisory and, the RAG augmented answer focusing on nutrient management of Citrus fruits. }
\label{fig:india-example}
\end{listing}
\section{Metrics}\label{sec:metrics}

The primary purpose of this section is to establish a comprehensive set of metrics designed to guide the quality assessment of the Q\&A generation process, particularly for the fine-tuning and RAG methodologies. A fundamental principle in machine learning is that the quality of the model is strongly influenced by the quality of the dataset it is trained on. In this context, to train or fine-tune a model that effectively generates high-quality Q\&A pairs, we first need a robust dataset. The inputs of our framework is the Q\&A pairs along with their associated context, and the output a quantification the quality of these pairs and, finally, of the fine-tuned model. The quality here is determined on a set of parameters. 

In the development of our metrics, several key factors must be taken into account. Firstly, the subjectivity inherent in question quality presents a significant challenge. Since opinions vary on what constitutes a relevant, informative, or engaging question, it's crucial to devise metrics that can objectively evaluate quality amidst this subjectivity. Secondly, the metrics must consider the dependency of a question's relevance and usefulness on its context. A question that provides valuable insights in one context may be deemed irrelevant in another, underscoring the need for context-aware metrics.

Thirdly, we need to evaluate the diversity and novelty of the generated questions. A robust question generation system should produce a wide array of questions covering various aspects of the given content. However, quantifying diversity and novelty can be challenging, as it involves assessing the questions' uniqueness and their similarity to the content and other questions generated. Grammatical correctness and fluency are also vital elements of question quality. While automated tools can evaluate grammatical correctness, assessing fluency often necessitates human judgment, making it difficult to create a fully automated metric. Lastly, a good question should be answerable based on the provided content. Evaluating whether a question can be accurately answered using the available information requires a deep comprehension of the content and the ability to identify relevant information that addresses the question.

In the existing literature, it's common to find a wealth of metrics designed to evaluate the quality of answers. These metrics play an indispensable role in ensuring that the answers provided by models are accurate, relevant, and effectively address the questions posed. However, there is a notable gap when it comes to metrics specifically designed to assess the quality of the questions themselves. Recognizing this lack, we have developed our metrics with a focus on evaluating question quality. Given the pivotal role that questions play in driving meaningful conversations and generating useful answers, it is equally important to ensure their quality as it is for the answers. This quantitative evaluation of the Q\&A sets will not only help in refining the generation process but also ensure the relevance and accuracy of the output, ultimately enhancing the overall efficacy of our model. Our developed metrics aim to bridge this gap in the literature, offering a comprehensive means of assessing question quality, which, we believe, will significantly contribute to the advancement of the Q\&A generation process.

\subsection{Question Evaluation}\label{sec:questionevaluation}

Defining metrics to evaluate the quality of questions generated by large language models (LLMs) is a challenging task due to several factors. The inherent complexity of natural language and the subjective nature of what constitutes a "good" question make it difficult to establish a universally accepted set of metrics.

To overcome these challenges, researchers often resort to a combination of automated and human evaluation methods. Automated methods, such as ROUGE~\cite{lin-2004-rouge} or BLEU~\cite{papineni-etal-2002-bleu} scores, can provide some insights into the quality of the generated questions, while human evaluation can help assess more subjective aspects like relevance, novelty, and fluency. Nevertheless, defining a comprehensive set of metrics for evaluating the quality of questions generated by LLMs remains an open and complex problem in the field of natural language processing. The proposed metrics are designed to provide a comprehensive view of the performance and effectiveness of a Q\&A generation system, allowing for targeted improvements and refinements and to filter the questions with low quality. The metrics we have developed to evaluate the questions are as follows: 

\begin{table}
\centering    
\begin{tabular}{>{\arraybackslash}p{6.8cm}>{\centering\arraybackslash}p{1.5cm}>{\arraybackslash}p{6.8cm}} 
\toprule  
\textbf{Question} & \textbf{Relevance Score $\uparrow$} & \textbf{Explanation} \\   
\midrule  
How can adaptive management practices benefit soil health in the inland Pacific Northwest? &  5 & The context heavily focuses on the challenges faced by farmers in the inland Pacific Northwest region and various strategies that can be adopted to overcome those challenges. The farmer would likely be interested in understanding how they can manage and improve soil health in the region, given the challenges of soil erosion, soil organic matter depletion, and consequent soil fertility loss, and the book covers this topic in detail in Chapter 2. Therefore, this question would be highly relevant and informative for a farmer in the region. \\ \hline  
How can understanding insect pests' distribution, life cycle, and crop damage help in their management in the inland Pacific Northwest?
 & 1 & The provided context does not mention anything about insect pests, so it is unlikely that a farmer would ask this question based on the given information. \\ \hline  
\end{tabular} \vspace{0.2cm} 
\caption{Examples of Relevance scores 1 and 5 for questions generated and present in the WA state benchmark dataset.} 
\label{table:q_relevance_metric}  
\end{table} 

\begin{table}
\centering    
\begin{tabular}{>{\arraybackslash}p{6.8cm}>{\centering\arraybackslash}p{1.5cm}>{\arraybackslash}p{6.8cm}} 
\toprule  
\textbf{Question} & \textbf{Global Relev. Score $\uparrow$} & \textbf{Explanation} \\   
\midrule  
How can diversifying or intensifying cropping systems benefit growers in the inland Pacific Northwest?&  5 &This question is highly informative and is likely to be asked by a farmer in the inland Pacific Northwest region who wants to optimize their crop yields and profitability. By asking this question, the farmer wants to understand how diversifying or intensifying their cropping systems can benefit them in the long run.  \\ \hline  

What is the purpose of downscaling global climate models (GCMs) in regional studies? &  1 & While climate change affects farming, this question is too technical in nature and not directly related to farming activities. It is unlikely that a farmer would ask this question. \\ \hline  
\end{tabular}\vspace{0.2cm}  
\caption{Examples of Global Relevance scores 1 and 5 for questions generated and present in the WA state benchmark dataset.}  
\label{table:glb_relevance_metric}  
\end{table}

\begin{itemize}
\item \textbf{Relevance}: To measure the informativeness of a generated Q\&A pair from the perspective of a farmer, we employ Large Language Model (LLMs) - namely GPT-4 - to rate the question on a scale of 1 to 5, with 5 being a question that would be asked by a farmer and 1 a question that would not, given the context. This metric ensures that the generated content is relevant and accurate to the target audience, considering all provided information. An example is provided in Table~\ref{table:q_relevance_metric}.

\item \textbf{Global Relevance}: To measure the informativeness of a generated Q\&A pair from the perspective of a farmer without considering any context, we employ Large Language Model (LLMs) - namely GPT-4 - to rate the question on a scale of 1 to 5, with 5 being a question that would be asked by a farmer and 1 a question that would not. This metric ensures that the generated content is relevant and accurate to the target audience. An example is provided in Table~\ref{table:glb_relevance_metric}.

\item \textbf{Coverage}: To gauge whether the generated answers can be directly extracted from the provided context and how well the model avoids hallucinating, we use LLMs to rate the answerability of each Q\&A pair on a scale of 1 to 5. The prompt was the following: “Your task is to rate from 1 to 5 if the answer can be extracted from the context and the question”. A higher score indicates that the answer can be more reliably extracted from the context, ensuring the model's output is grounded in the available information. An example is provided in Table~\ref{table:q_coverage_metric}.

\item \textbf{Overlap}: To assess the semantic similarity between the generated question and the source text/section, we use Kullback-Leibler (KL) divergence as a measure. KL divergence is an asymmetric measure that quantifies the difference between two probability distributions, in this case, the word probability distributions of the source text and the concatenated questions. By leveraging the smoothed counts of words in the texts, this method allows for a more nuanced comparison of their semantic content. A lower KL divergence value indicates a higher degree of semantic overlap between the questions and the text, which is desired for effective Q\&A generation.
 
\item \textbf{Diversity}: We also aim to evaluate the variety of questions generated by the system. To do this, we Word Mover's Distance (WMD) as a measure. Word Mover's Distance (WMD) is a powerful metric for measuring the semantic similarity between two documents or text samples~\citep{kusner}. It is based on the concept of optimal transport, which seeks to find the most efficient way to "move" the content of one document to another in terms of the semantic meaning of their words. WMD leverages word embeddings, such as Word2Vec or GloVe, which represent words as high-dimensional vectors capturing their semantic meaning. These embeddings enable the computation of distances between words in a continuous semantic space, allowing for a more nuanced comparison of the meanings of two text samples. A smaller WMD value indicates a higher degree of overlap in meaning between the question and the text, which is desired for effective Q\&A generation. We then compute a similarity matrix between all generated questions. By calculating the average of the off-diagonal elements in this matrix, we can obtain a measure of diversity. A smaller average value indicates greater similarity among questions and, therefore, less diversity. 

\item \textbf{Details}: We assess the level of detail in both the generated questions and answers by counting the number of tokens (words) in each. This metric provides insight into the depth and specificity of the content generated by the Q\&A system. By employing these metrics, we can effectively evaluate and refine the Q\&A generation process, ensuring that the generated content is informative, relevant, diverse, and grounded in the source material. This will ultimately lead to a more useful and effective Q\&A generation system for the target audience. 

\item \textbf{Fluency}: Finally, we use the fluency metric that leverages large language models, such as GPT-4, to assess fluency and coherence. By preparing a prompt that instructs the AI to rate a given question on a scale of 1 to 5 and provide an explanation, the model can generate insightful scores for each question. After appending the generated question to the prompt, submitting it to GPT-4, and parsing the response to extract the fluency score and explanation, the results can be stored for further analysis. This approach effectively utilizes GPT-4’s language understanding capabilities to assess the quality of generated questions, aiding in refining question-generation models and selecting the best questions for specific applications. An example is provided in Table~\ref{table:fluency_metric}.

\end{itemize}

\begin{table}
\centering    
\begin{tabular}{>{\arraybackslash}p{6.8cm}>{\centering\arraybackslash}p{1.5cm}>{\arraybackslash}p{6.8cm}} 
\toprule  
\textbf{Question} & \textbf{Coverage Score $\uparrow$} & \textbf{Explanation} \\   
\midrule  
What are the potential impacts of climate change on dryland wheat production systems in the inland Pacific Northwest? &  5 & The context provides specific information on climate change impacts on dryland wheat production systems in the inland Pacific Northwest, including temperature-induced drought conditions and increased frequency and severity of extreme weather events. The context also notes how climate change may exacerbate historical conditions linked to soil erosion events in the region. \\   
\midrule  
How does the topography of the Pacific Northwest influence the region's climate and agricultural zones? &  1 & The context provides information about wheat production, climate change, soil health, tillage systems, crop residue management, rotational diversification, soil fertility, soil amendments, precision agriculture, weed management, and disease management in the region, but it does not directly address the influence of the topography on the climate and agricultural zones. \\   
\bottomrule  
\end{tabular}\vspace{0.2cm}    
\caption{Examples of Coverage scores 1 and 5 for questions generated and present in the WA state benchmark dataset.} 
\label{table:q_coverage_metric}    
\end{table}

\begin{table}
\centering    
\begin{tabular}{>{\arraybackslash}p{6.8cm}>{\centering\arraybackslash}p{1.5cm}>{\arraybackslash}p{6.8cm}} 
\toprule  
\textbf{Question} & \textbf{Fluency Score $\uparrow$} & \textbf{Explanation} \\   
\midrule  
How do wind and water erosion impact the cereal-producing regions of the Inland Pacific Northwest? &  5 &The question is fluent, coherent, and makes sense. It is a clear and specific question that is directly related to the concerns and interests of farmers who produce cereals in the Inland Pacific Northwest. It demonstrates an understanding of the potential threats posed by wind and water erosion to crop production and suggests a desire to learn more about these issues in order to mitigate their impact on agricultural productivity. \\  
\midrule
Why has the reliability of the Haney's test been questioned? &  3 & The question is coherent and makes sense, but it lacks context. It is unclear what Haney's test refers to and what the question is trying to achieve.\\   
\bottomrule  
\end{tabular} \vspace{0.2cm}   
\caption{Examples of Fluency scores 3 and 5 for questions generated and present in the WA state benchmark dataset. One should note that almost no values below 3 were obtained.} 
\label{table:fluency_metric}    
\end{table}

\subsection{Answer Evaluation}

Evaluating answers generated by LLMs is challenging due to their tendency of generating long, informative, and conversational answers. Traditional metrics are unsuitable to evaluate this type of answers~\citep{kamalloo_evaluating_2023, adlakha_evaluating_2023}. Recent works have shown that LLMs themselves are judges that display high agreement with humans~\citep{chiang_can_2023, dubois_alpacafarm_2023} and can be used directly or in conjunction with other techniques to evaluate answers to questions with or without context~\citep{min_factscore_2023}. There is also promise in fine-tuning LLMs as judges for open-ended scenarios~\citep{zhu_judgelm_2023}. In this work, we have used the AzureML Model Evaluation~\citep{azuremlmetrics}, with the following metrics to compare the generated answers with ground truth:

\begin{itemize}

\item \textbf{Coherence}: Comparison of coherence between ground truths and predictions given the context. The metric provides a score between one to five, where one means that the answer lacks coherence and five means the answer has perfect coherency. An example is provided in Table~\ref{table:a_coherence_metric}. 

\renewcommand{\arraystretch}{1.5}
\begin{table}
\centering    
\begin{tabular}{p{4cm}p{2cm}p{9cm}}    
\hline
\textbf{Question} & \textbf{Coherence Score $\uparrow$} & \textbf{Answer} \\
\hline
What are the negative effects of wind erosion on soil? & 5 & Wind erosion can have several negative effects on soil, including: 1) Loss of topsoil: Wind erosion can lead to the loss of topsoil, which can negatively impact crop productivity and soil fertility. 2) Airborne pollutants: Wind-blown dust and soil particles can be considered airborne pollutants and can pose a human health hazard. 3) Decreased water quality: Eroded soil can be deposited into surface water, affecting water quality and potentially harming aquatic life. 4) Reduced soil organic matter: Wind erosion can lead to the loss of soil organic matter, which can negatively impact soil fertility and structure. 5) Negative impact on root zone and seedbed environments: Wind erosion can reduce the water storage potential of soil and negatively influence root zone and seedbed environments, leading to reduced crop productivity.\\ 
\textit{Expert answer} & & Here are five key negative effects of wind erosion on soil: 1) Nutrient Loss: Wind erosion often removes the nutrient-rich topsoil, which is vital for plant growth, leading to poorer crop yields.
2) Soil Degradation: The removal of fine particles and organic matter leaves behind a coarser soil, less able to retain water and therefore less fertile. 3) Environmental Impact: Dust and particles carried away by wind erosion can degrade air quality, affecting human, animal, and plant health. 3) Ecosystem Disruption: The physical changes caused by wind erosion can disrupt ecosystems and lead to habitat loss for some species. 5) Economic Costs: The consequences of wind erosion, such as reduced agricultural productivity and health impacts, can have significant economic costs.\\  \hline
What is the effect of ensiling on the levels of nitrates and sulfur-containing compounds in canola? & 1 & Ensiling canola increases the levels of nitrates and sulfur-containing compounds. \\ 
\textit{Expert answer} & & Ensiling can reduce both nitrate levels and sulfur-containing compounds in canola. This happens as microorganisms during the fermentation process convert nitrates into proteins and break down some sulfur compounds. The extent of reduction can vary based on specific ensiling conditions.\\ 
\\\hline    
\end{tabular}\vspace{0.2cm}    
\caption{Examples of Coherence scores 1 and 5 for answers generated using a fine-tuned \texttt{Llama2-13b model} and the expert answer for each question.}  
\label{table:a_coherence_metric}    
\end{table} 

\item \textbf{Relevance}: Relevance measures how well the answer addresses the main aspects of the question based on the context. The metric rates from 1 to 5, where 5 means the answer has perfect relevance. An example is provided in Table~\ref{table:a_relevance_metric}.

\begin{table}
\centering    
\begin{tabular}{>{\arraybackslash}p{6.8cm}>{\centering\arraybackslash}p{1.5cm}>{\arraybackslash}p{6.8cm}} 
\toprule  
\textbf{Question} & \textbf{Relevance Score $\uparrow$} & \textbf{Answer} \\   
\midrule  
 What is the purpose of collecting soil compaction data?&  5 & The purpose of collecting soil compaction data is to determine the extent of soil compaction, which can help farmers make informed decisions about soil management practices and improve long-term soil health.\\   
\midrule  
What is the only species of spider documented as venomous to humans in the Pacific Northwest? &  1 &Based on the information provided, the only species of spider documented as venomous to humans in the Pacific Northwest is the Pacific Rattlesnake, \textit{Crotalus oreganus oreganus}.\\  
\bottomrule  
\end{tabular}\vspace{0.2cm}  
\caption{Examples of Relevance scores 1 and 5 for answers generated using a fine-tuned \texttt{Llama2-13b model}.} 
\label{table:a_relevance_metric}    
\end{table}  

\item \textbf{Groundedness}: The metric defines weather the answer follows logically from the information contained in the context or not and provides and integer score to determine how grounded the answer is. An example is provided in Table~\ref{table:a_groundedness_metric}.

\begin{table}
\centering    
\begin{tabular}{>{\arraybackslash}p{6cm}>{\centering\arraybackslash}p{2.3cm}>{\arraybackslash}p{6.8cm}} 
\toprule  
\textbf{Question} & \textbf{Groundedness Score $\uparrow$} & \textbf{Answer} \\   
\midrule  
 What are some potential impacts of weed control on wildlife in Christmas tree farms? &  5 & Weed control in Christmas tree farms can have both positive and negative impacts on wildlife. On the one hand, weed control can improve the overall health and vigor of the trees, leading to increased survival rates and better tree-to-land ratio. This can have a positive impact on wildlife, as it can provide more habitat and food resources for various species. However, the use of herbicides for weed control can also have negative impacts on wildlife. For example, the loss of weeds can lead to a decrease in food resources for some species, such as birds and small mammals, which may rely on weeds for food and shelter. Additionally, the use of herbicides can have unintended consequences, such as harming beneficial insects or other wildlife species. It is important to consider the potential impacts of weed control on wildlife when making management decisions in Christmas tree farms. Integrated pest management (IPM) strategies that include a combination of cultural, biological, and chemical controls can help minimize the risks to wildlife while achieving effective weed control.\\ 
\midrule  
What are some benefits of quinoa?
 &  1 & Quinoa is a good source of protein, fiber, minerals, and B vitamins. It is gluten-free and can be used as a substitute for rice or couscous. It is also high in antioxidants and has been shown to lower cholesterol and blood sugar levels. Additionally, quinoa is a good source of essential fatty acids and has been used to help people lose weight.\\  
\bottomrule  
\end{tabular}\vspace{0.2cm}    
\caption{Examples of Groundedness scores 1 and 5 for answers generated using a fine-tuned \texttt{Llama2-13b model}.} 
\label{table:a_groundedness_metric}    
\end{table}

\item \textbf{Completion}: Number of tokens in each answer.

\end{itemize}

\subsection{Model Evaluation}\label{sec:model-evaluation}

Evaluation of fine-tuned LLM’s poses a significant challenge, as human evaluation is expensive, and the correctness of technical answers can be difficult for non-experts to determine. To evaluate the different fine-tuned models, we used GPT-4 as an evaluator. GPT-4 was used to generate approximately 270 question and answer pairs from the agriculture documents as the ground truth dataset. Answers for each FT and RAG model were generated for these questions. To ensure the consistency of GPT-4 as an evaluator, we prompted GPT-4 five times per evaluation to measure the variance in responses. The LLMs were evaluated across a number of different metrics: 

\begin{itemize}

\item \textbf{Evaluation with Guideline}: For each Q\&A ground truth pair, GPT-4 was prompted to generate an evaluation guideline, to list what a correct answer should contain. Then, GPT-4 was prompted to score each answer from 0 to 1, based on whether it fulfilled the criteria in the evaluation guideline. One example is shown bellow:

\begin{listing}
    \begin{mdframed}[linewidth=1pt,linecolor=black,backgroundcolor=gray!20]
  \centering
\begin{lstlisting}
Question: What is the life cycle of the Cherry Bark Tortrix? 

Answer: The Cherry Bark Tortrix produces one prolonged generation per year, with the adult flight period extending from May to September, often with two flight periods. It survives the winter as larvae feeding beneath the bark until spring.

Evaluation_guideline: The answer should mention that the Cherry Bark Tortrix has one prolonged generation per year, with an adult flight period from May to September, and that it survives the winter as larvae feeding beneath the bark.
\end{lstlisting} 
\end{mdframed}
  \caption{Question, answer and evaluation guideline used.}
  \label{lst:example-guideline}
\end{listing}

\item \textbf{Succinctness}: We created a scoring sheet describing what succinct and verbose answers might contain. We prompted GPT-4 with the scoring sheet, the ground truth answer, and the LLM answer, and asked for a grade on a scale from 1 to 5. An example is provided in Table~\ref{table:SuccinctnessExamples}.

\item \textbf{Correctness}: We created a scoring sheet describing what a complete, partially correct, or incorrect answer should contain. We prompted GPT-4 with the scoring sheet, the ground truth answer, and the LLM answer, and asked for a grade of correct, incorrect, partially correct. We provide two examples in Tables~\ref{table:PartiallyCorrectAnswer} and~\ref{table:FullyCorrectAnswer}.

\end{itemize}

\begin{table}
\centering
\begin{tabular}{p{3cm}p{12cm}}    
\hline
\textbf{Succinctness $\uparrow$} &  \\
\hline
\textbf{Score} & \textbf{5} \\
Question & What is the scientific name of the white pine weevil? \\ 
Answer & The scientific name of the white pine weevil is Pissodes strobi. \\ 
Reference Answer &  Pissodes strobi \\ 
Explanation & The response is on point and does not contain any additional information. The user is able to understand the point quickly. \\  
\hline
\textbf{Score} & \textbf{2} \\
Question & What is the purpose of salting fish before smoking? \\ 
Answer & The purpose of salting fish before smoking is to help remove moisture from the fish and to preserve it by inhibiting the growth of bacteria. Salt draws out the moisture from the surface of the fish, making it firm and creating a barrier that prevents bacteria from multiplying. It also enhances the flavor of the fish and helps to prevent it from spoiling. This process is commonly used in curing and preserving fish, and is an important step in the preparation of smoked fish.  \\ 
Reference Answer &  Salting fish before smoking preserves the fish by reducing its moisture content.  \\ 
Explanation & The predicted answer provides a lot of additional information about the process of salting fish, including how it works and why it's important. While this information is relevant and informative, it goes beyond the scope of the original question, which simply asked for the purpose of salting fish before smoking. Therefore, the response is verbose.  \\  
\hline
\end{tabular}\vspace{0.2cm}    
\caption{Examples of Succinctness scores for GPT-4 generated answers.}
\label{table:SuccinctnessExamples}    
\end{table} 

\begin{table}
\centering    
\footnotesize
\begin{tabular}{p{3cm}p{12cm}}    
\hline
\textbf{Correctness} &  \\
\hline
\textbf{Score} & \textbf{Partially Correct} \\
Question & What are some factors to consider when focusing on soil health testing in vineyards?  \\ 
Answer & Based on the context information, some factors to consider when focusing on soil health testing in vineyards include:
\begin{itemize}
\item Consistency with the timing of sampling and the soil testing lab.
\item Understanding the interpretation of the measured value from the lab.
\item Focusing testing on functions you want to improve.
\item Keeping good records of lab results and vineyard management for long-term soil health tracking.
\item Not expecting changes to happen immediately.
\item Assessing block variability for effective management.
\item Having fun exploring soil through a new lens.
\end{itemize}
It's important to note that these factors are not mutually exclusive and may overlap or interact with each other in different ways. Additionally, the specific factors that are most important may vary depending on the specific vineyard and its management goals.  \\ 
Reference Answer &  Some factors to consider when focusing on soil health testing in vineyards include the functions you want your soil to serve (e.g., supplying plant nutrients, reducing erosion, or increasing aggregate stability), the cost of soil health testing, the frequency of sampling for different soil properties, and the time it takes to see measurable differences after a change in management.  \\ 
Explanation & The answer mentions several factors to consider when focusing on soil health testing in vineyards, such as consistency with the timing of sampling, understanding the interpretation of the measured value, focusing on functions to improve, keeping good records, not expecting immediate changes, assessing block variability, and having fun exploring soil. Although it does not directly mention the cost of testing and the frequency of sampling, it covers other relevant factors and provides a comprehensive response.  \\  
\hline
\end{tabular}\vspace{0.2cm}    
\caption{Example of a partially correct answer. Although the answer contains many factors related to soil health testing, it does not mention the cost of testing, an important factor.}
\label{table:PartiallyCorrectAnswer}    
\end{table} 

\begin{table}
\centering 
\footnotesize   
\begin{tabular}{p{3cm}p{12cm}}    
\hline
\textbf{Correctness} &  \\
\hline
\textbf{Score} & \textbf{Correct} \\
Question & What is the concern with using herbicide-resistant wheat technology to control jointed goatgrass?  \\ 
Answer & The concern with using herbicide-resistant wheat technology to control jointed goatgrass is that the genetic similarity between jointed goatgrass and wheat allows hybridization between the two species, with the potential for movement of the resistance gene into a jointed goatgrass population, resulting in herbicide-resistant jointed goatgrass. This can lead to the development of herbicide-resistant weeds, which can reduce the effectiveness of herbicides and increase the risk of weed resistance. Additionally, the use of herbicide-resistant wheat technology can lead to the overuse of herbicides, which can have negative environmental impacts and contribute to the development of herbicide-resistant weeds.   \\ 
Reference Answer &  The concern with using herbicide-resistant wheat technology to control jointed goatgrass is that the resistance genes could move from wheat into jointed goatgrass populations, potentially creating herbicide-resistant jointed goatgrass.   \\ 
Explanation & The answer correctly identifies the concern with using herbicide-resistant wheat technology to control jointed goatgrass, which is the potential for the resistance gene to move into a jointed goatgrass population, resulting in herbicide-resistant jointed goatgrass. The answer also provides additional information about the potential negative impacts of using herbicide-resistant wheat technology, such as the overuse of herbicides and the development of herbicide-resistant weeds. This additional information does not detract from the answer and provides a more comprehensive response to the question.   \\  
\hline
\end{tabular}\vspace{0.2cm}    
\caption{Example of fully correct answer. The answer contains all of the necessary information in the response, as well as some additional relevant information, and is therefore rated as correct.}
\label{table:FullyCorrectAnswer}    
\end{table}

\section{Experiments}\label{sec:experiments}

The primary objective of the experiments section is to offer a comprehensive evaluation and analysis of various models and their performance in generating Q\&A pairs within the context of agricultural data and using these for RAG or fine-tuning. This assessment is crucial as it provides insight into the performance of these models and their potential applications in various settings. It further aids in identifying and understanding the strengths and weaknesses of different models and techniques, thereby offering a foundation for future developments.

The experiments section is structured into several self-contained experiments, each focusing on a specific aspect of Q\&A generation and evaluation, RAG and fine-tuning. These experiments are designed to be independent yet complementary, providing a multifaceted understanding of the performance of the models.

The experiments explore the following areas:
\begin{itemize}
\item \textbf{Q\&A Quality:} This experiment assesses the quality of the generated Q\&A pairs by three Large Language Models (LLMs), namely GPT-3, GPT-3.5, and GPT-4, under different context setups. The quality is evaluated based on several metrics including relevance, coverage, overlap, and diversity.
\item \textbf{Context Study:} This experiment investigates the impact of different context setups on the performance of the models in generating Q\&A pairs. It evaluates the generated pairs under three context setups: no context, context, and external context.
\item \textbf{Model to Metrics Calculation:} This experiment compares how GPT-3.5 and GPT-4 behave when used to calculate the metrics for assessing the quality of the Q\&A pairs.
\item \textbf{Combined vs Separated Generation:} This experiment explores the benefits of generating questions and answers separately as opposed to generating them together, focusing on the efficiency in token usage.
\item \textbf{Retrieval Ablation Study:} This experiment evaluates the retrieval capabilities of Retrieval-Augmented Generation (RAG), a method that enhances the inherent knowledge of LLMs by providing additional context during question answering.
\item \textbf{Fine-tuning:} This experiment evaluates the performance of fine-tuned models versus base instruction-tuned models. It aims to understand the potential of fine-tuning to help the models learn new knowledge.
\end{itemize}

In summary, each experiment sheds light on different aspects of the pipeline, offering a holistic understanding of the performance of these models.

\subsection{Q\&A Quality}

In the following experiments, we employed three LLMs, namely GPT-3, GPT-3.5 and GPT-4, to evaluate the quality of generated Q\&A pairs. The pairs were generated under different contexts: No context, Context, and External context as explained in the next subsection \ref{contextstudy}. We used several metrics (defined on Section~\ref{sec:questionevaluation}) to assess their quality, including Relevance, Global Relevance, Coverage, Overlap, Diversity, Details, and Fluency. The LLMs assessed the Q\&A pairs on various metrics, scoring each on a scale from 1 to 5. For certain metrics such as Overlap and Diversity, we incorporated intricate methods like Kullback-Leibler (KL) divergence and Word Mover's Distance (WMD) to measure the semantic similarity between the source text and the questions generated. The primary objective of this experimental setup was to comprehend how different Q\&A generation techniques can influence the quality of the resulting Q\&A pairs. We aimed to discern how various metrics perceive and gauge the quality of these pairs and to highlight any significant disparities in their evaluations. This insight is crucial for refining the Q\&A generation process, as it ensures that the resulting content is informative, pertinent, diverse, and fluent.

\subsubsection{Context Study}\label{contextstudy}

GPT-3, GPT-3.5, and GPT-4 models were tasked to generate Q\&A pairs using three different context setups: \textit{no context}, \textit{context}, and \textit{external context}. The \textit{no context} setup provided no specific information in the prompt. The \textit{context} setup included the location or state from where the document originated in the prompt. In the \textit{external context} setup, additional context obtained through the process outlined in Section~\ref{sec:augmentation} was included in the prompt. An example if provided in Table~\ref{tab:context}. The performance of these models in generating Q\&A pairs was assessed using various metrics as presented in Table~\ref{table:questionevaluation}. These metrics provided insights into the coverage, size of the prompt, diversity, overlap, relevance, and fluency of the generated Q\&A pairs across the different context setups.

\begin{table}
\centering    
\begin{tabular}{>{\arraybackslash}p{3cm}>{\arraybackslash}p{13cm}} 
\toprule  
\textbf{Context} &  \textbf{Question} \\   
\midrule  
No context &  What are some decision support tools that can help producers in the Pacific Northwest understand the impacts of climate change and variability on their operations? \\  
\midrule
Context &  As a dryland farmer in the Inland Pacific Northwest, USA, how can decision support tools help in adapting to climate change and its impacts on agricultural operations?\\ 
\midrule
External context & What decision support tools are available to agricultural decision-makers in the Inland Pacific Northwest to help them understand how climate change and variability may affect their operations?\\  
\bottomrule  
\end{tabular}  \vspace{0.2cm}  
\caption{Examples of questions generated with \textit{no context}, \textit{context} and \textit{external context}.} 
\label{tab:context}    
\end{table}

\begin{table}
\centering  
\footnotesize 
\begin{tabular}{llccccccc}  
\hline  
\textbf{GPT version} & \textbf{Context} & \textbf{Coverage $\uparrow$} & \textbf{Prompt Size} & \textbf{Diversity $\downarrow$} & \textbf{Overlap $\uparrow$} & \textbf{Relevance $\uparrow$} & \textbf{Fluency $\uparrow$} \\    
\hline  
3 & No context & 3.67 & 12.05 & 1.16 &  \textcolor[HTML]{239f3f}{1.04} & 4.74 & 4.78 \\  
3.5 & No context & 3.90 & 12.50 & 1.14 & 1.02 & 4.70 & 4.70 \\  
4 & No context & \textcolor[HTML]{239f3f}{3.97} & 14 & 1.13 & 0.98 & 4.80 & 4.80 \\  
3 & Context & 3.18 & 18.33 &  \textcolor[HTML]{239f3f}{0.92} & 1.02 & 4.74 & 4.72 \\  
3.5 & Context & 3.78 & 16.99 & 1.06 & 0.98 & 4.73 & 4.73 \\  
4 & Context & 3.89 & 18.52 & 0.96 & 0.95 & 4.77 & 4.77 \\  
3 & External context & 3.21 & 17.77 &0.93 &  \textcolor[HTML]{239f3f}{1.04} & 4.75 & 4.77 \\  
3.5 & External context & 3.83 & 16.36 & 1.06 & 0.98 & 4.75 & 4.75 \\  
4 & External context & \textcolor[HTML]{239f3f}{3.97} & 19.44 & \textcolor[HTML]{239f3f}{0.93} & 0.96 &  \textcolor[HTML]{239f3f}{4.81} &  \textcolor[HTML]{239f3f}{4.81} \\  
\hline  
\end{tabular}\vspace{0.2cm}  
\caption{Metrics calculated for different combinations of context and model. For almost all metrics, higher values are desirable, but for diversity, where smaller values indicate a more diverse set of questions.}  
\label{table:questionevaluation}  
\end{table}

In the \textit{no context} setup, GPT-4 has the highest Coverage and Size Prompt among the three models, suggesting that it can cover more text sections but it generates more verbous questions. However, all three models have similar values for Diversity, Overlap, Relevance, and Fluency. When the context is included, GPT-3.5 shows a slight increase in Coverage compared to GPT-3, while GPT-4 maintains the highest value of Coverage. The Size Prompt is the largest for GPT-4, indicating its capacity to generate more verbose questions and answers. In terms of Diversity and Overlap, all three models show similar performance. For Relevance and Fluency, there's a slight increase in GPT-4 compared to the other models.

In the \textit{external context} setup, similar trends are observed. GPT-4 has the highest Coverage and Size Prompt, while all models show similar Diversity and Overlap. The  Relevance and Fluency are slightly higher for GPT-4. In summary, GPT-4 generally shows superior performance in terms of Coverage and Size Prompt across all context setups. The three models show similar performance in terms of Diversity, Overlap, while GPT-4 has a slight edge in terms of Relevance and Fluency.

Moreover, looking at each model, the \textit{no context} setup seems to provide the best balance for GPT-4 in terms of average coverage, diversity, overlap, relevance and fluency, but generates shorter Q\&A pairs. The \textit{context} setup leads to longer Q\&As and a slight decrease in most metrics, except for the size. The \textit{external context} setup, while generating the longest Q\&A, maintains the average coverage and sees a slight increase in average relevance and fluency. 

In summary, the \textit{no context} setup seems to provide the best balance for GPT-4 in terms of average coverage, diversity, overlap, relevance and fluency, but generates shorter answers. The \textit{context} setup leads to longer prompts and a slight decrease in the other metrics. The \textit{external context} setup, while generating the longest prompts, maintains the average coverage and sees a slight increase in average relevance and fluency. Therefore, the choice between the three would depend on the specific requirements of your task. If the length of the prompts is not being taken into consideration, the \textit{external context} could be the best choice due to the higher relevance and fluency scores. 

In the different context setups, GPT-3 shows varying performance. The \textit{no context} setup yields the best balance in terms of average coverage, diversity, overlap, relevance, and fluency, but it generates the shortest prompts. When context is included, there is a decrease in most metrics except for the size of the prompt, which increases. The \textit{external context} setup shows a slight improvement in coverage and diversity compared to the \textit{context} setup and maintains the stability of overlap, relevance, and fluency. However, it generates slightly shorter prompts than the \textit{context} setup but longer than the \textit{no context} setup. Overall, GPT-3 performs best in a \textit{no context} setup when considering all metrics but generates shorter prompts.

For GPT-3.5, the \textit{no context} setup results in reasonable performance in terms of diversity, overlap, relevance, and fluency, with the average size prompt being relatively shorter. When context is included, there is an increase in average coverage and diversity, while the average size prompt decreases. The \textit{external context} setup sees a slight increase in average coverage, relevance, and fluency compared to the \textit{context} setup, with the average size prompt decreasing further. While the \textit{context} setup has the highest average coverage and diversity, the \textit{external context} setup sees slight increases in average relevance and fluency. The choice between these setups for GPT-3.5 would depend on the specific requirements, with the \textit{external context} setup potentially being the best choice if higher relevance and fluency scores are preferred.

\subsubsection{Model to Metrics Calculation}

Here, we compared how GPT-3.5 and GPT-4 behave when used to calculate the metrics presented on Section~\ref{sec:questionevaluation}. The results are shown in Table~\ref{table:metriccomparison}. Here, all questions and answers were generated in combination using GPT-3. 

\begin{table}
\centering  
\footnotesize 
\begin{tabular}{llcccccc}  
\hline  
\textbf{Evaluation} & \textbf{Context} & \textbf{Coverage $\uparrow$} &\textbf{ Prompt Size }& \textbf{Diversity $\downarrow$} & \textbf{Overlap $\uparrow$} & \textbf{Relevance $\uparrow$} & \textbf{Fluency $\uparrow$} \\    
\hline  
GPT-3.5 & No context & 3.67 & 12.05 & 1.16 &  \textcolor[HTML]{239f3f}{1.04} & 4.74 & 4.78 \\  
GPT-3.5 & Context & 3.18 & 18.33 & 0.92 & 1.02 & 4.74 & 4.72 \\  
GPT-3.5 & External context & 3.21 & 17.77 & 0.93 &  \textcolor[HTML]{239f3f}{1.04} &  \textcolor[HTML]{239f3f}{4.75} & 4.77 \\  
GPT-4 & No context & 4.34 & 13.14 &  \textcolor[HTML]{239f3f}{0.74} & 0.80 & 4.12 & 4.81 \\  
GPT-4  & Context & 4.30 & 15.52 &  \textcolor[HTML]{239f3f}{0.74} & 0.80 & 4.06 &  \textcolor[HTML]{239f3f}{4.84} \\  
GPT-4  & External context &  \textcolor[HTML]{239f3f}{4.50} & 15.64 & 0.77 & 0.81 & 4.20 & 4.83 \\  
\hline  
\end{tabular}\vspace{0.2cm}  
\caption{Metrics comparison when calculated using GPT-3.5 and GPT-4.}  
\label{table:metriccomparison}  
\end{table}

In terms of Coverage, which gauges how well the answer can be extracted from the provided context, the GPT-4 model gives higher ratings across all contexts compared to the GPT-3.5 model. This suggests that GPT-4 may perceive the generated Q\&A pairs as more grounded in the available context than GPT-3.5.

When it comes to Diversity, a measure of the variety of generated questions, GPT-4 provides lower scores across all contexts. This could indicate that GPT-4 perceives less semantic variation among the generated questions compared to GPT-3.5. Looking at the Overlap metric, which assesses the semantic similarity between the generated question and the source text, GPT-4 gives lower scores across all contexts. This suggests that GPT-4 perceives less semantic overlap between the questions and the source text, implying more unique question generation.

In terms of Relevance, a measure of the informativeness of a generated Q\&A pair from a farmer's perspective, GPT-4 gives lower scores across all contexts compared to GPT-3.5. This could suggest that GPT-4 perceives the generated Q\&A pairs as less relevant to a farmer's perspective than GPT-3.5. Finally, the Fluency metric, which assesses the fluency and coherence of the generated text, shows that GPT-4 gives higher scores across all contexts. This suggests that GPT-4 rates the generated questions as more fluent and natural-sounding compared to GPT-3.5.

In conclusion, while GPT-4 generally rates the generated Q\&A pairs as more fluent and grounded in context, it also perceives them as less diverse and relevant compared to the ratings by GPT-3.5. These insights are crucial for understanding how different models perceive and evaluate the quality of generated content. 

However, when considering the two models, GPT-3.5 and GPT-4, one of the most noticeable differences is their performance speed. GPT-3.5 tends to be faster compared to GPT-4. For example, when using Azure OpenAI, \texttt{gpt-35-turbo} and \texttt{gpt-35-turbo-16k} generate between 240k and 300k tokens per minute depending on the region. On the other hand, \texttt{GPT-4} and \texttt{GPT-4-32k} generate between 20k and 80k depending on the region and model. Considering this trade-off, the majority of the metrics for Q\&A generation were calculated using GPT-3.5, while the model evaluation metrics were calculated using GPT-4.

\subsubsection{Combined vs Separated Generation}
In another experiment, we explored the benefits of generating questions and answers separately as opposed to generating them together. The primary advantage of this approach is the efficiency in token usage. By generating questions and answers separately, fewer tokens are used, which can be particularly beneficial when working within the token limits of language models.

Moreover, generating questions and answers separately opens up the possibility of using different approaches or models for each component of the Q\&A pair. For instance, one could utilize a Retrieval-Augmented Generation (RAG) model or a fine-tuned language model for answering, thereby not being solely dependent on base language models. This flexibility allows for a more tailored and potentially effective approach to Q\&A generation. The metrics calculate are presented on Table~\ref{table:combinedseparated}. The pairs or questions were generated using GPT-4 and the metrics were calculated using GPT-3.5.

\begin{table}
\centering  
\begin{tabular}{llccccc}  
\hline  
\textbf{Generation Method} & \textbf{Context} & \textbf{Coverage $\uparrow$} &  \textbf{Diversity $\downarrow$} & \textbf{Overlap $\uparrow$} & \textbf{Relevance $\uparrow$}  & \textbf{Fluency $\uparrow$}\\    
\hline  
Combined generation & No context & 3.97 & 1.13 & \textcolor[HTML]{239f3f}{0.98} & 4.80 & 4.80\\  
Combined generation & Context & 3.89 & 0.96 & 0.95 & 4.77 & 4.77\\  
Combined generation & External context & 3.97 & 0.93 & 0.96 & \textcolor[HTML]{239f3f}{4.81}& 4.81 \\  
Only questions & No context & 4.34 & \textcolor[HTML]{239f3f}{ 0.74} & 0.80 & 4.12 & 4.81\\  
Only questions & Context & 4.30 &  \textcolor[HTML]{239f3f}{0.74} & 0.80 & 4.06 & \textcolor[HTML]{239f3f}{4.84}\\  
Only questions & External context & \textcolor[HTML]{239f3f}{4.50} &  0.77 & 0.81 & 4.20& 4.83 \\  
\hline  
\end{tabular}\vspace{0.2cm}  
\caption{Metrics generated for different combinations of context and generation method.}  
\label{table:combinedseparated}  
\end{table}  

When assessing Coverage, which measures how well the generated content covers the text used, the \textit{only questions} method consistently performs better than \textit{combined generation} across all contexts. This suggests that when the task is to generate \textit{only questions} from the text, the model is able to cover more ground. The \textit{combined generation} method generally has higher Diversity scores compared to the \textit{only questions} method, indicating that it generates less varied content. 

Overlap, which measures the similarity between the generated content and the source material, is a little higher for the \textit{combined generation} method across all contexts. This suggests that the \textit{combined generation} method produces content that is more consistent with the source material. However, that could be due to the fact that the generated content is longer and, because of that, more similar to the text used.

In terms of Relevance, the \textit{combined generation} method outperforms the \textit{only questions} method in all contexts. This suggests that generating both questions and answers produces content that is more relevant to the context, when considering a farmer perspective. On the other hand, we tested changing the prompt used to calculate Relevance and removed the farmer perspective. With that, the score of both setups was extremely similar. Lastly, for Fluency, which measures the readability and naturalness of the generated content, both methods perform similarly across all contexts, indicating that both methods generate highly fluent content.

In summary, while the \textit{only questions} method provides better coverage and desirable lower diversity, the \textit{combined generation} method scores higher in terms of overlap and relevance. Both methods perform similarly in terms of fluency. Therefore, the choice between these two methods would depend on the specific requirements of the task. If the goal is to cover more information and maintain more diversity, the \textit{only questions} method would be preferred. However, if maintaining close overlap with the source material, the \textit{combined generation} method would be the better choice.

\subsection{Retrieval Ablation Study}

Retrieval-Augmented Generation (RAG) enhances the inherent knowledge of LLMs by providing additional context during question answering. Especially in the agricultural context, these auxiliary snippets of information are crucial to reduce hallucinations and tailor the answer to the geographic region or phenomenon of interest to the farmer. To properly augment the LLM’s capabilities, RAG must be able to retrieve the relevant snippets from the corpus of supporting documents. In this section, we perform an ablation study on RAG’s retrieval capabilities considering the agricultural dataset explored in this work.

We measure recall to understand how well RAG retrieves the original content used to generate a question in our dataset. The questions were generated based on the 573 documents of the Washington state benchmark dataset, and, for each question, we stored the passage of text used as basis for its creation. Following the RAG setup detailed in Section~\ref{sec:methodology_answer_generation_rag}, our goal is to retrieve snippets that are contained in the original excerpt for a given question. In this context, recall refers to how many times we are able to retrieve the original excerpt given a question. We aim for higher recall as LLMs showed to be able to filter out unrelated pieces of information and highlight essential bits as long as the correct context is present during the answer generation.

Initially, we investigate the impact of the number of snippets retrieved (i.e., top-$k$) and present the results in Table~\ref{table:rag_ablation_snippets}. By considering more snippets, RAG is able to recover the original excerpt more consistently. Additionally, as we split the documents into chunks when building the FAISS index, snippets may represent distinct parts of the original document, covering complementary information that can be useful during answer generation. With this in mind, we consider $k = 3$ snippets in the following experiments of this work, as it is able to recover the original passage more than $80\%$ of times without considerably increasing the input prompt size.

To ensure that the model can handle questions from various geographical contexts and phenomena, we might need to expand the corpus of supporting documents to cover a wide range of topics. As we consider more documents, we anticipate the index will grow in size. This could potentially increase the number of collisions among similar snippets during the retrieval process, hindering our ability to recover the relevant information for the input question, thus decreasing our recall. To evaluate this behavior, we incrementally rebuild the FAISS index, adding documents from other states besides Washington. We compute recall@top-3 on the Washington benchmark dataset for each setup and report the results in Table~\ref{table:rag_ablation_index_size}. With more documents, recall decreases as collisions become more common. Despite that, the relevant context is retrieved more than $75\%$ of cases even when the index increases sixfold (from 573 to 3888 documents).

\begin{table}[t]
\begin{minipage}{.4\linewidth}
\centering
\begin{tabular}{lc}
\hline
\textbf{Top-$k$} & \textbf{Recall} \\  
\hline
1 & $61.9\%$  \\
2 & $75.0\%$  \\
3 & $81.5\%$  \\
5 & $87.8\%$  \\
10 & $93.6\%$ \\
\hline
\end{tabular}\vspace{0.2cm}
\captionof{table}{Recall at different number of retrieved snippets (top-$k$).}
\label{table:rag_ablation_snippets}
\end{minipage}\hspace{.1\linewidth}
\begin{minipage}{.5\linewidth}
\centering
\begin{tabular}{lc}
\hline
\textbf{\# of documents} & \textbf{Recall@top-3}\\  
\hline
573 & $81.5\%$  \\
847 & $79.5\%$  \\
1249 & $78.1\%$ \\
2460 & $76.0\%$  \\
3888 & $75.0\%$  \\
\hline
\end{tabular}\vspace{0.2cm}
\captionof{table}{Recall as the index size grows in number of documents.}
\label{table:rag_ablation_index_size}
\end{minipage}
\end{table}

\subsection{Fine-tuning}
In this section, we evaluate the performance of fine-tuned models versus base instruction-tuned models.
The fine-tuned models are trained on a dataset of questions extracted from the Washington state dataset, a comprehensive collection of 573 documents entailing approximately 2 million tokens.
The answers for the questions were generated from the Llama2-13B-chat model with RAG.
The evaluation dataset is composed of 273 human-curated for the state of Washington.
Each sample in the evaluation dataset consists of a question and an evaluation guideline that describes the contents of a desirable answer.
We generated answers for both base and fine-tuned models with and without RAG, and used the metrics discussed in \ref{sec:model-evaluation} to evaluate the answers.

For base models, we evaluated the open source models Llama2-13B-chat and Vicuna-13B-v1.5-16k.
Both models are relatively small, and represent an interesting trade-off of compute vs performance.
Both models are fine-tuned version of Llama2-13B using different methods.
Llama2-13B-chat was instruction-tuned via supervised fine-tuning and reinforcement learning (RLHF)~\cite{ouyang2022rlhf, touvron_llama_2023-1}.
Vicuna-13B-v1.5-16k is instruction-tuned via supervised fine-tuning on the ShareGPT dataset~\cite{zheng_judging_2023}.
Additionally, we also evaluated base GPT-4, as a larger, more expensive and more capable alternative.
For fine-tuned models, we fine-tune Llama2-13B directly on agriculture data, in order to compare its performance against similar models tuned for more generic tasks.
We also fine-tune GPT-4, in order to evaluate if fine-tuning can still be helpful on very large models.

\subsubsection{Evaluation with Guideline}

The results for \emph{Evaluation with Guideline} are shown in Table~\ref{table:fine-tuned}.
In the baseline scenario, GPT-4 emerged as the top performer when integrated with RAG, securing a score of 80\%. Vicuna and Llama2-chat 13B trailed closely, scoring 79\% and 75\%, respectively. It is noteworthy that the results of GPT-4 and Llama2-chat 13B without the integration of RAG were quite similar. When it came to the fine-tuned models, GPT-4 displayed exceptional performance, securing an accuracy of 86\% with RAG and 81\% without RAG. These findings underscore the consistent superior performance of GPT-4 across all scenarios.

To further illustrate the performance of these models, we have provided in Table~\ref{table:examples} a few examples of a single question and the diverse responses generated by different models. These models include the reference answer model, Vicuna, Vicuna integrated with RAG, GPT-4, and GPT-4 post-fine-tuning. The variation in answers from these different models serves to contextualize and elaborate on the performance metrics presented in Table \ref{table:examples}.

This table presents a comparative analysis of responses generated by different AI models to a specific question about the main concern of blowing dust from excessively tilled fallow fields in the low precipitation wheat production region of the Inland Pacific Northwest. The reference answer, serving as the benchmark, is "Soil loss and air quality."

Starting with the Vicuna + RAG model, it offers a succinct response that aligns well with the reference answer, addressing both soil loss and air quality issues. This model is one of the best performers in terms of directly addressing the reference answer in a concise manner. The Llama2 13B + RAG and Llama2 13B fine-tuned models provide responses that are relevant, but they do not directly address soil loss, which is a significant part of the reference answer. The Llama2 13B fine-tuned + RAG model addresses both air quality and health issues related to PM10 emissions, but it doesn't explicitly mention soil loss.

GPT-4's response is the most comprehensive, covering a wide array of concerns. However, its answer might be seen as too extensive for those seeking a more direct answer. The GPT-4 + RAG model provides a clear and concise response that directly addresses the reference answer. It highlights soil loss, reduced air quality, and the health risks associated with inhaling dust, making it one of the best performers. The GPT-4 fine-tuned model offers an in-depth response that focuses on soil loss and indirectly addresses air quality through mentions of air pollution and climate change. Its answer, while comprehensive, aligns well with the reference answer, making it another top performer. Finally, the GPT-4 fine-tuned + RAG model provides a detailed answer that directly addresses both soil loss and air quality concerns. It also provides additional context about the characteristics of the soil and emissions in the region, making it one of the most accurate and informative models.

\begin{table}
\centering
\begin{tabular}{lcccccccc}
\hline
\textbf{Model} & \multicolumn{1}{c}{\textbf{Fine-tuned}} & \multicolumn{1}{c}{\textbf{Accuracy}} & \multicolumn{1}{c}{\textbf{+RAG}} & \\  
\hline
Llama-2-chat 13B & &76\% $ \pm 2 \%  $  & 75\% $ \pm 2 \%  $    \\
Vicuna & &72\% $ \pm 2 \%  $ & 79\% $ \pm 2 \%  $ &     \\
GPT-4 & &75\% $ \pm 3 \%  $ &80\% $ \pm 4 \%  $ &    \\
Llama2 13B & $\checkmark$ &68\% $ \pm 3 \%  $ & 77\% $ \pm 2 \%  $ &     \\
GPT-4 & $\checkmark$& 81\%$ \pm 5 \%  $  & 86\% $ \pm 2 \%  $ &    \\
\hline
\end{tabular}\vspace{0.2cm}
\caption{Accuracy of base and fine-tuned models with and without RAG.}
\label{table:fine-tuned}
\end{table}

\subsubsection{Succinctness}

To comprehensively measure the quality of the responses, we evaluated responses for succinctness in addition to the accuracy. As shown above, responses can be excessively long and detract from the quality of the responses. Across the models, RAG integration leads to more succinctness of responses (Table \ref{table:succintness}, as the responses are typically focused on the context provided. GPT-4+RAG tends to provide the most succinct responses, with a score of 3.2 out of 5, though we do not see this improvement with fine-tuned GPT-4+RAG. Depending on user preference or the specific use case, more or less verbose answers may be preferred. This should be considered in addition to accuracy, especially for models demonstrating similar performance.

\begin{table}
\centering
\begin{tabular}{lcccccccc}
\hline
\textbf{Model} & \multicolumn{1}{c}{\textbf{Fine-tuned}} & \multicolumn{1}{c}{\textbf{Succintness $\uparrow$}} & \multicolumn{1}{c}{\textbf{+RAG}} & \\  
\hline
Llama-2-chat 13B & & 2.1  & 2.8  &   \\
Vicuna & & 2.7  & 3.1    & \\
GPT-4 & & 2.3  & 3.2   & \\
Llama2 13B & $\checkmark$ & 2.7 &  2.8   \\
GPT-4 & $\checkmark$ & 2.5 &  2.5   \\
\hline
\end{tabular}\vspace{0.2cm}
\caption{Succinctness of base and fine-tuned models with and without RAG, on a scale from 1 (verbose) to 5 (succinct).}
\label{table:succintness}
\end{table}

\subsubsection{Correctness}

The examples in Table~\ref{table:examples} show that these models do not consistently provide a full answer to the question. For example, some responses note soil loss but not air quality as a concern (or vice versa). We therefore also prompted GPT-4 to evaluate the answers on whether they were fully correct. Similar to the valuation with a guideline, we see that GPT-4 and LLama-2-chat 13B outperform Vicuna. With the addition of RAG, we see a large increase in the percent of answers that are fully correct for all models. As with succinctness, we observe that RAG focuses responses on the most relevant subset of information.  We see the biggest improvement for Vicuna with the addition of RAG, going from 28\% to 56\% fully correct. Fine-tuned GPT-4 outperforms the rest of the models with and without RAG, with 45\% and 61\% of responses fully correct, respectively.

\begin{table}
\centering
\begin{tabular}{lcccccccc}
\hline
\textbf{Model} & \multicolumn{1}{c}{\textbf{Fine-tuned}} & \multicolumn{1}{c}{\textbf{Fully correct (\%)}} & \multicolumn{1}{c}{\textbf{+RAG}} & \\  
\hline
Llama-2-chat 13B    & & 32\%  & 49\%  &   \\
Vicuna              & & 28\%  & 56\%    & \\
GPT-4                & & 36\%  & 60\%   & \\
Llama2 13B  & $\checkmark$ & 29\% &  49\%   \\
GPT-4        & $\checkmark$ & 45\% &  61\%   \\
\hline
\end{tabular}\vspace{0.2cm}
\caption{Percent of answers that were fully correct, for base and fine-tuned models with and without RAG.}
\label{table:correctness}
\end{table}

In conclusion, in this example, the best performing models in terms of accurately and succinctly addressing the reference answer are Vicuna + RAG, GPT-4 + RAG, GPT-4 fine-tuned, and GPT-4 fine-tuned + RAG. These models provide a balanced mix of precision, conciseness, and depth of information.

\begin{table}
\centering
\footnotesize
\begin{tabular}{|p{3cm}|p{12cm}|}  
\hline  
\textbf{Reference Question} & \textbf{What is the main concern of blowing dust from excessively tilled fallow fields in the low precipitation wheat production region of the Inland Pacific Northwest?} \\  
\hline  
Reference Answer & Soil loss and air quality.\\  
\hline  
Vicuna & The main concern of blowing dust from excessively tilled fallow fields in the low precipitation wheat production region of the Inland Pacific Northwest is soil erosion and loss of topsoil. The region has a semi-arid climate with low precipitation, which makes it prone to wind erosion. When fields are tilled excessively, the soil becomes loose and can be easily blown away by strong winds, leading to a loss of topsoil and a reduction in soil fertility. This can have a negative impact on crop yields and the long-term productivity of the land. Additionally, blowing dust can also pose health risks to people living in the area, as it can contain harmful pollutants and particles.\\  
\hline  
Vicuna + RAG & The main concern of blowing dust from excessively tilled fallow fields in the low precipitation wheat production region of the Inland Pacific Northwest is the reduced air quality and increased soil losses due to the repeated tillage practices.\\  
\hline   
Llama2 13B + RAG & Based on the information provided, the main concern of blowing dust from excessively tilled fallow fields in the low precipitation wheat production region of the Inland Pacific Northwest is the negative impact on air quality due to the inhalation of the dust into lung tissue.\\ 
\hline 
Llama2 13B fine-tuned & Based on the information provided, the main concern of blowing dust from excessively tilled fallow fields in the low precipitation wheat production region of the Inland Pacific Northwest is soil erosion.\\ 
\hline 
Llama2 13B fine-tuned + RAG & Based on the context information provided, the main concern of blowing dust from excessively tilled fallow fields in the low precipitation wheat production region of the Inland Pacific Northwest is reduced air quality and respiratory health issues due to PM10 emissions.\\  
\hline  
GPT-4 & The main concern of blowing dust from excessively tilled fallow fields in the low precipitation wheat production region of the Inland Pacific Northwest is its potential negative impact on air quality, human health, crop yields, soil erosion, and ecosystem health. The excessive tillage of fallow fields can lead to the loss of topsoil, organic matter, and nutrients, making the soil more vulnerable to erosion and dust generation. The blown dust can reduce visibility, increase respiratory problems, and deposit harmful pollutants, such as herbicides and pesticides, far away from their target areas. The dust can also damage crops by smothering the leaf surface, reducing photosynthesis, and increasing water stress. Moreover, the eroded soil and dust can affect the water quality of nearby streams and rivers by increasing sedimentation and nutrient loading. Therefore, it is crucial to adopt sustainable and innovative farming practices, such as conservation tillage, crop rotation, cover crops, and windbreaks, to reduce the negative impacts of dust and erosion while maintaining crop productivity.\\  
\hline  
GPT-4 + RAG & The main concern of blowing dust from excessively tilled fallow fields in the low precipitation wheat production region of the Inland Pacific Northwest is that it leads to repeated soil losses, reduced air quality, and poses a health risk when inhaled into lung tissue.\\
\hline  
GPT-4 fine-tuned & The main concern of blowing dust from excessively tilled fallow fields in the low precipitation wheat production region of the Inland Pacific Northwest is the loss of soil organic matter and nutrients, which can lead to soil degradation and reduced crop yields. Additionally, blowing dust can cause air pollution and contribute to climate change. It is important to implement conservation tillage practices, such as no-till or reduced-till farming, to minimize soil disturbance and reduce the risk of blowing dust.\\  
\hline  
GPT-4 fine-tuned + RAG & Based on the information provided, the main concern of blowing dust from excessively tilled fallow fields in the low precipitation wheat production region of the Inland Pacific Northwest is the loss of soil and reduced air quality. The text states that blowing dust leads to repeated soil losses and reduced air quality, and that wind erosion and dust emissions mostly occur in low precipitation areas with sandy silt loam soils that are poorly aggregated and dominated by particulates 100 micrometers in diameter. Additionally, the text notes that excessive tillage during summer fallow pulverizes soil clods and buries residue, which can contribute to soil loss and reduced air quality.\\  
\hline 
\end{tabular}\vspace{0.2cm}  
\caption{Comparison of answers provided by different models for a reference question.}  
\label{table:examples}  
\end{table}

\subsection{Knowledge Discovery}

Our study aimed to explore the potential of fine-tuning to help GPT-4 learn new knowledge, which is crucial for applied research. To test this, we selected questions across the 50 states in the US that were similar across at least three states. We then computed the cosine similarity from the embeddings and identified a list of 1000 such questions. These questions were removed from the training set, and we used both fine-tuning and fine-tuning with RAG to assess whether GPT-4 could learn new knowledge based on similarities across different states.

The results from our study, as shown in Table \ref{table:newknowledge}, were quite promising. The data revealed that GPT-4 was only able to learn 47\% of the new knowledge presented to it. However, with the help of fine-tuning, we were able to significantly increase this percentage. Specifically, the fine-tuned model was able to learn up to 72\% and 74\% of the new knowledge, depending on whether we used RAG or not. This upper bound represents the maximum amount of new knowledge that the model could have learned, given the data it was presented with.

It is worth noting that the potential applications of this research are numerous. For instance, consider a scenario where we need to evaluate the impact of a particular pest or disease on corn. In some cases, we may not have access to advisory information specific to our region. However, we may have access to advisory information from a similar region, which could still be useful. By leveraging fine-tuning and RAG, we can help the model learn this new knowledge and apply it to our specific situation.

\begin{table}
\centering
\begin{tabular}{lcccccccc}
\hline
\textbf{Model} & \multicolumn{1}{c}{\textbf{GPT-4}} & \multicolumn{1}{c}{\textbf{Fine-tuned}} & \multicolumn{1}{c}{\textbf{Fine-tuned+RAG}} & \\  
\hline
Similar & 47\% &72\%   & 74\%    \\
Somewhat similar & 49\% & 27\%  & 25\%  &     \\
Not similar & 4\% &1\%  &1\%  &    \\
\hline
\end{tabular}\vspace{0.2cm}
\caption{Accuracy of the base models and the fine-tuned models with and without RAG.}
\label{table:newknowledge}
\end{table}

The results from Table \ref{table:newknowledge} demonstrate that GPT-4 learned just 47\% of this new knowledge and with fine-tuning we were able to get this number way higher to 72\% and 74\% with RAG and the fine-tuned model. Interestingly, this can be defined as the upper bound of what the model could have learned.

\section{Conclusion}\label{sec:conclusion}

This study aimed to establish a baseline for assessing the capabilities of large
language models, such as LLama 2, GPT-3.5 and GPT-4, in addressing complex
problems in agriculture. By evaluating their performance when using RAG or/and fine-tuning, the study provides valuable insights into the strengths and
limitations of LLMs within the agricultural domain. 

The primary contributions of the paper include the establishment of performance baselines for LLMs when using RAG or fine-tuning as it presents different benefits and costs. RAG, known for improving accuracy in large models, is highly effective in instances where data is contextually relevant, such as in the interpretation of farm data. The low initial cost of creating embeddings, which are vector representations of data, makes RAG an attractive option. However, it is important to consider that the input token size can increase the prompt size, and the output token size tends to be more verbose and harder to steer.

On the other hand, fine-tuning offers a precise, succinct output that is attuned to brevity. It is highly effective and presents opportunities to learn new skills in a specific domain, like improving crop yield predictions or optimizing irrigation schedules based on weather patterns. However, the initial cost is high due to the extensive work required to fine-tune the model on new data. In addition, fine-tuning necessitates minimal input token size, making it a more efficient option for handling large data sets. A comparison between two approaches is presented on Table~\ref{table:Conclusion}.

\begin{table}
\centering
\begin{tabular}{lcccccccc}
\hline
\textbf{Model} & \multicolumn{1}{c}{\textbf{RAG}}& \multicolumn{1}{c}{\textbf{Fine-tuning}} \\  
\hline
Cost – input token size & Increased Prompt Size & Minimal\\
Cost – output token size & More verbose, harder to steer  & Precise, tuned for brevity\\
Initial cost & Low – creating embeddings & High – fine-tuning\\
Accuracy & Effective & Effective \\
New Knowledge & If data is in context & New skill in domain\\
\hline
\end{tabular}\vspace{0.2cm}
\caption{Insights on RAG vs Fine-tunning.}
\label{table:Conclusion}
\end{table}

This research also serves as a pioneering step in establishing a pipeline for the application of RAG and fine-tuning techniques in various LLMs, enabling innovation and collaboration across multiple industries. With our initial focus on agriculture, we have demonstrated how these strategies can lead to more efficient models starting from the Q\&A generation process.

In this study, we also demonstrated how to generate relevant question and answers for datasets for specific industries, by leveraging structured document understanding, together with GPT-4 for question generation, and RAG for answer generation. The questions generated were highly specific to the respective sections they were derived from, and the model was able to utilize the entire text for generating insightful and comprehensive answers. Our exploration indicated that generating questions and answers separately leads to efficient token usage, opening up the possibility of using different models or approaches for each component of the Q\&A pair. We also proposed a series of metrics to properly evaluate the quality of the generated questions relative to the information contained in the original documents, and showcased multiple metrics for measuring the quality of the RAG-generated answers. 

As we progress, it is essential to continually refine our comprehension of different LLMs' capabilities, including RAG and fine-tuning. Even though GPT-4 consistently outperformed other models, the costs associated with its fine-tuning and inference cannot be ignored and are an important trade off to consider. In conclusion, while both RAG and fine-tuning are effective techniques, their suitability would depend on the specific application, the nature and size of the data set, and the resources available for model development. Nevertheless, this work paves the way for further investigations on how to best combine the two approaches and further explorations on dataset generation pipelines for industry-specific LLM applications. As future work, further investigation of the kind of knowledge the fine-tuned model gains is going to be important, more investigation in how to improve structured extraction from the documents and leverage when developing systems using LLMs. Another exciting direction is on how to combine the structured information from PDFs with images and captions from the same documents to enable multi-modal fine-tuning opportunities.

\newpage
\bibliographystyle{plainnat}
\bibliography{paper}

\end{document}